\documentclass[12pt]{article}

\usepackage{tikz}
\tikzset{font={\fontsize{8pt}{10}\selectfont}}

\usepackage{geometry}
\usetikzlibrary{calc}
\usetikzlibrary{arrows,shapes,chains}
\usepackage{xcolor}
\usepackage{amsfonts}	
\usepackage{amsmath}
\usepackage{bm}
\usepackage{graphicx}
\usepackage{multirow}
\usepackage{algorithmic}
\usepackage{algorithm}
\usepackage{subfigure}
\usepackage{epstopdf}
\usepackage{url}
\usepackage{cleveref}
\usepackage{cases}
\crefname{equation}{}{}
\crefname{table}{TABLE}{TABLES}
\crefname{figure}{Fig.}{Figs.}
\crefname{section}{Section}{Sections}

\DeclareMathOperator*{\argmax}{arg\,max}

\newcommand{\cb}[1]{\ifmmode {\boldsymbol{#1}}\else ${\boldsymbol{#1}}$\fi}

\newcommand{\cp}[1]{\ifmmode {\mathcal{#1}}\else ${\mathcal{#1}}$\fi}

\begin{document}

% \begin{frontmatter}



\title{Improving Deep Hyperspectral Image Classification Performance with Spectral Unmixing}
\date{}


% \author[label1]{Alan~J.X.~Guo}
% \author[label1]{Fei~Zhu}
% %\author[label1]{Fei~Zhu\corref{cor1}}
% %\cortext[cor1]{Corresponding author \ead {fei.zhu@tju.edu.cn}}

% \address{
% \address[label1]{Center for Applied Mathematics, Tianjin University, China.}
% %\address[label3]{Corresponding author.}
% }
\author{Alan~J.X.~Guo and Fei~Zhu\thanks{A.~Guo, and F.~Zhu are with the Center for Applied Mathematics, Tianjin University, China. (jiaxiang.guo;~fei.zhu@tju.edu.cn) }
}

\maketitle

\begin{abstract}
Recent advances in neural networks have made great progress 
	in the hyperspectral image (HSI) classification. 
	However, the overfitting effect, 
	which is mainly caused by complicated 
	model structure and small training set, 
	remains a major concern.
	Reducing the complexity of 
	the neural networks could prevent overfitting to some extent, 
	but also declines the networks' ability to express more abstract features. 
	Enlarging the training set is also difficult, 
	for the high expense of acquisition and manual labeling. 
	In this paper, 
	we propose an abundance-based multi-HSI classification method. 
	Firstly, we convert every HSI from the spectral domain to the abundance domain 
	by a dataset-specific autoencoder. 
	Secondly, the abundance representations 
	from multiple HSIs are collected to form an enlarged dataset. 
	Lastly, we train an abundance-based classifier 
	and employ the classifier to predict over all the involved HSI datasets.
	Different from the spectra that are usually highly mixed, the abundance features are 
	more representative in reduced dimension with less noise. 
	This benefits the proposed method to employ simple classifiers and enlarged training data, 
	and to expect less overfitting issues.
	The effectiveness of the 
	proposed method is verified by the ablation study and the comparative experiments. 
	% On four datasets, the proposed method provides comparable results with two state-of-the-art methods 
	% by a much simpler model. 

\end{abstract}

% \begin{keyword}

% Convolutional neural networks \sep autoencoder \sep spectral unmixing \sep hyperspectral image classification

% \end{keyword}

% \end{frontmatter}

\section{Introduction}
Hyperspectral image (HSI) is a data cube consisting of
reflection or radiance spectra, acquired by the remote sensors when 
flying over real-world objects or scenes. The height and width of an HSI are 
decided by the monitored scene at a specific resolution, while the depth 
records the measurements across a certain wavelength range. Thus, 
each pixel corresponds to a spectral vector. 
In the past decades, the HSI analysis has witnessed rapid development with 
plenty of applications~\cite{bioucas2012hyperspectral, Chang2003Hyperspectral}. 
To effectively explore the rich spectral and spatial information contained in HSIs, 
different categories of processing techniques have been proposed, {\em e.g.}, 
the spectral unmixing~\cite{bioucas2012hyperspectral}, 
the classification~\cite{wang2016survey}, 
the image restoration~\cite{Fan2017Hyperspectral},
and the target detection~\cite{NasrabadiHyperspectral}. 
In this paper, we particularly address the deep learning-based HSI classification problem, 
where the spectral unmixing technique is smartly integrated to the classification task, such that the classification model 
is simplified and the training set is amplified. This helps to improve the 
classification performance.

The HSI classification refers to classifying each of the pixels 
to a certain class according to the spectral and spatial characteristics. 
Earlier approaches were developed based on conventional machine learning algorithms, 
including the principal component analysis (PCA) \cite{jiang2018superpca}, 
the independent component analysis (ICA) \cite{villa2011hyperspectral},
the linear discriminant analysis (LDA) \cite{li2011locality},
the support vector machine (SVM) \cite{melgani2004classification}, 
and the sparse representation \cite{Fang2014Spectral}, 
to name a few. 
Despite the great progress recently brought by the neural networks (NN) to HSI classification, 
the none-NN methods continue to play a role independently or 
as a part of the NN-based algorithms \cite{neware2018survey,pachon2018random,huang2020local,chang2019statistical}, 
on account of the drawbacks of the NN, 
{\em e.g.}, the overfitting problem on small training sets \cite{paoletti2019deep}.

Spectral unmixing (SU) is another active area of research in HSI analysis. 
It is assumed that each spectrum is a mixture of several ``pure'' material signatures, termed endmembers. 
The aim of SU is to extract the endmembers and to estimate their respective proportions, 
namely the abundance fractions, at each pixel~\cite{keshava2002spectral}. 
Recently, several works have introduced the SU as a complementary source of 
information in HSI classification. 
In \cite{dopido2011unmixing,dopido2012quantitative}, the SU was applied to 
reduce the spectral dimension, in order to avoid the Hughes phenomenon when applying 
the SVM classifier. 
In \cite{ibarrola2019hyperspectral}, the authors investigated several SU methods and assigned the label to a pixel 
according to its maximum abundance.
In \cite{villa2011spectral,fang2020combining}, the extracted abundances 
were used as supplementary information to improve the classification 
accuracy on the hard samples, namely the highly-mixed spectra.
The authors in \cite{alam2017combining} considered a region-based nonnegative 
matrix factorization for band group based abundance estimation. 
The abundance matrices at different ranges of wavelengths were used as input 
to train a convolutional neural network (CNN) based classifier. 
Moreover, in the scope of HSI classification by semisupervised learning, 
the SU was applied in selecting the most informative samples 
\cite{dopido2014new,li2015complementarity,samat2016improved,sun2016new}. 

The NN has gained great popularity and achieved 
remarkable results in many machine learning fields, 
{\em e.g.,} the computer vision, especially after the introduction of CNN and deep learning \cite{lecun2015deep, krizhevsky2012imagenet}. 
Since then, numerous investigations have been made to apply deep learning-based algorithms to 
the HSI analysis. 
In the \cite{cai2020bsnet,roy2020darecnetbs}, the authors used NN-based model to conduct band selection tasks.
The researchers of \cite{wang2019nonlinear, ZhaoHyperspectral} applied the NN-based autoencoder for SU on HSIs.
In the field of HSI classificaiton, many of the early works considered to utilize the stacked autoencoder (SAE) 
to extract denoised or sparse features 
from the spectral or spatial-spectral data, 
and the obtained features were 
usually classified with a traditional 
classifier, such as the SVM and the logistic regression 
\cite{chen2014deep, ma2016spectral,tao2015unsupervised,zhou2019learning,zhang2017recursive}. 
More recently, several end-to-end classification methods based on CNN 
\cite{slavkovikj2015hyperspectral,chen2016deep,jiao2017deep,paoletti2020rotation,yao2020clustercnn}
and the recurrent neural network (RNN) \cite{mou2017deep,zhang2018spatial} 
have been proposed and significantly improved the classification results. 
See \cite{paoletti2019deep} for an overview of the deep learning methods for HSI classification. 

To train a deep learning-based model typically 
requires a large amount of labeled data, 
otherwise, the learned model would be prone to overfitting. 
However, the availability of training samples is limited in HSIs due to the 
high expense of acquisition and manual labeling~\cite{chen2016deep}. 
To tackle this contradiction, different strategies have proved their effectiveness 
in existing works, {\em e.g.}, data augmentation and transfer learning. 
In \cite{chen2016deep}, a virtual sample enhanced method was proposed to
improve the performance of the CNN-based model. 
In \cite{li2017hyperspectral}, the authors designed a pixel-pair-based model, 
where the training set is composed of pixel-pairs instead of pixels.
This ensures the sufficiency of labeled samples for training a deep CNN.
Alternatively, transfer learning has been employed to alleviate the overfitting issue, 
that is, by transferring the knowledge acquired from the source domain to the target domain, 
the demand of training samples would be reduced.
Knowledge was transferred from the ordinary 
RGB images to HSI classification tasks~\cite{jiao2017deep}, and transferred from multiple HSIs
to the classification tasks on small-scale HSIs \cite{zhao2020classification}. 
Moreover, the authors in \cite{xing2016stacked} applied the knowledge learned from unsupervised 
tasks to classification tasks on the same HSI, by transferring a pre-trained 
stacked denoising autoencoder and fine-tuning on the labeled samples. 

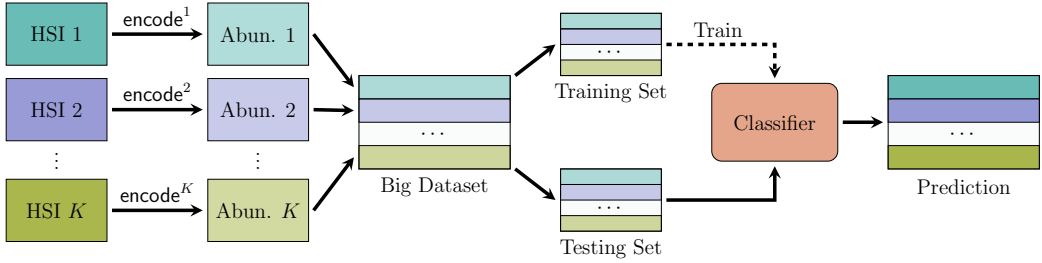
\begin{figure*}
	\centering
{\linespread{1}
	\centering
	\tikzstyle{format}=[circle,draw,thin,fill=white]
	\tikzstyle{format_gray}=[circle,draw,thin,fill=gray]
	\tikzstyle{format_rect}=[rectangle,draw,thin,fill=white,align=center]
	\tikzstyle{arrowstyle} = [->,thick]
	\tikzstyle{network} = [rectangle, minimum width = 3cm, minimum height = 1cm, text centered, draw = black,align=center,rounded corners,fill=green_so,fill opacity=0.5,text opacity=1]
	\tikzstyle{training_batch} = [trapezium, trapezium left angle = 30, trapezium right angle = 150, minimum width = 3cm, text centered, draw = black, fill = cyan_so, fill opacity=0.3,text opacity=1,align=center]		
	\tikzstyle{class_features} = [trapezium, trapezium left angle = 30, trapezium right angle = 150, minimum width = 3cm, text centered, draw = black, fill = cyan_so, fill opacity=0.3,text opacity=1,align=center]
	\tikzstyle{pixel} = [rectangle, draw = black, fill = orange_so, fill opacity=0.5,text opacity=0,align=center]	
	\tikzstyle{pixel_red} = [rectangle, draw = black, fill = red_so, fill opacity=1,text opacity=0,align=center]	
	\tikzstyle{feature} = [rectangle, draw = black, fill = orange_so, fill opacity=0.3,text opacity=0,align=center,rounded corners]	
	\tikzstyle{feature_sfp} = [rectangle, draw = black, fill = violet_so, fill opacity=0.3,text opacity=0,align=center,rounded corners]					
	\tikzstyle{arrow1} = [thick, ->, >= stealth]
	\tikzstyle{arrow1_thick} = [thick, ->, >= stealth, line width=2pt]
	\tikzstyle{arrow2} = [thick, dashed, ->, >= stealth]
	\tikzstyle{thick_line} = [line width=0.7pt,dashed]
	\tikzstyle{channel} = [fill=white,fill opacity = 0.9]
	\tikzstyle{channel_shadow} = [fill = gray_so, fill opacity = 0.1, rounded corners]
	\tikzstyle{channel_selected} = [fill = orange_so, fill opacity = 0.37]
	
	\begin{tikzpicture}[auto,>=latex',  thin,  start chain=going below, every join/.style={norm},font=\normalsize]
		\definecolor{gray_so}{RGB}{88,110,117}
		\definecolor{lightgray_so}{RGB}{207,221,221}
		\definecolor{yellow_so}{RGB}{181,137,0}
		\definecolor{cyan_so}{RGB}{42,161,152}
		\definecolor{orange_so}{RGB}{203,75,22}
		\definecolor{green_so}{RGB}{133,153,0}
		\definecolor{red_so}{RGB}{220,50,47}
		\definecolor{magenta_so}{RGB}{211,54,130}
		\definecolor{violet_so}{RGB}{108,113,196}
		% \useasboundingbox [fill=lightgray_so](0,0) rectangle (19.4*0.8,5.2*0.8);
		\useasboundingbox (0,0) rectangle (19.4*0.8,5.2*0.8);

		\scope[transform canvas={scale=0.67}]
		
		\coordinate (zero) at (0,0);
		\coordinate (HSIk) at ($(zero)+(0.2,0.2)$);
		\filldraw[channel_selected,fill=green_so,fill opacity = 0.7] ($(HSIk)$) rectangle ($(HSIk)+(2,2*0.618)$);
		\node at ($(HSIk)+(1,0.618)$) {{HSI $K$}};

		\node at ($(HSIk)+(1,0.7+1)$) {{$\vdots$}};

		\coordinate (HSI2) at ($(HSIk)+(0,2)$);
		\filldraw[channel_selected,fill=violet_so,fill opacity = 0.7] ($(HSI2)$) rectangle ($(HSI2)+(2,2*0.618)$);
		\node at ($(HSI2)+(1,0.618)$) {{HSI $2$}};

		\coordinate (HSI1) at ($(HSI2)+(0,1.5)$);
		\filldraw[channel_selected,fill=cyan_so,fill opacity = 0.7] ($(HSI1)$) rectangle ($(HSI1)+(2,2*0.618)$);
		\node at ($(HSI1)+(1,0.618)$) {{HSI $1$}};
		% hsi end abd begin
		\coordinate (ABDk) at ($(HSIk)+(4,0)$);
		\filldraw[channel_selected,fill=green_so] ($(ABDk)$) rectangle ($(ABDk)+(2,2*0.618)$);
		\node at ($(ABDk)+(1,0.618)$) {{Abun. $K$}};
		\draw [arrow1_thick] ($(HSIk)+(2.1,0.618)$) -- ($(ABDk)+(-0.1,0.618)$);
		\node at ($(ABDk)+(-1,0.97)$) { $\mathsf{encode}^K$};
		\node at ($(ABDk)+(1,0.7+1)$) {{$\vdots$}};

		\coordinate (ABD2) at ($(ABDk)+(0,2)$);
		\filldraw[channel_selected,fill=violet_so] ($(ABD2)$) rectangle ($(ABD2)+(2,2*0.618)$);
		\node at ($(ABD2)+(1,0.618)$) {{Abun. $2$}};
		\draw [arrow1_thick] ($(HSI2)+(2.1,0.618)$) -- ($(ABD2)+(-0.1,0.618)$);
		\node at ($(ABD2)+(-1,0.97)$) {$\mathsf{encode}^2$};

		\coordinate (ABD1) at ($(ABD2)+(0,1.5)$);
		\filldraw[channel_selected,fill=cyan_so] ($(ABD1)$) rectangle ($(ABD1)+(2,2*0.618)$);
		\node at ($(ABD1)+(1,0.618)$) {{Abun. $1$}};
		\draw [arrow1_thick] ($(HSI1)+(2.1,0.618)$) -- ($(ABD1)+(-0.1,0.618)$);
		\node at ($(ABD1)+(-1,0.97)$) {$\mathsf{encode}^1$};
		% abd end big set begin
		\coordinate (big) at ($(ABDk)+(3,3.5/2+0.618-0.618*1.5)$);
		\filldraw[channel,fill=lightgray_so,fill opacity = 0.1] ($(big)$) rectangle ($(big)+(3,3*0.618)$);
		\filldraw[channel_selected,fill=cyan_so] ($(big)+(0,2.25*0.618)$) rectangle ($(big)+(3,3*0.618)$);
		\filldraw[channel_selected,fill=violet_so] ($(big)+(0,1.5*0.618)$) rectangle ($(big)+(3,2.25*0.618)$);
		\filldraw[channel_selected,fill=green_so] ($(big)+(0,0*0.618)$) rectangle ($(big)+(3,0.75*0.618)$);
		\node at ($(big)+(1.5,-0.35)$) {{Big Dataset}};
		\node at ($(big)+(1.5,0.618*3*3/8)$) {$\cdots$};
		\draw [arrow1_thick] ($(ABDk)+(2.1,0.618)$) -- ($(big)+(-0.1,0.618*3*1/8)$);
		\draw [arrow1_thick] ($(ABD2)+(2.1,0.618)$) -- ($(big)+(-0.1,0.618*3*5/8)$);
		\draw [arrow1_thick] ($(ABD1)+(2.1,0.618)$) -- ($(big)+(-0.1,0.618*3*7/8)$);
		%big set end training testing begin
		\coordinate (training) at ($(big)+(4,0.618*3)$);
		% \filldraw[channel_selected,fill=yellow_so] ($(training)$) rectangle ($(training)+(2,2*0.618)$);
		\filldraw[channel,fill=lightgray_so,fill opacity = 0.1] ($(training)$) rectangle ($(training)+(2,2*0.618)$);
		\filldraw[channel_selected,fill=cyan_so] ($(training)+(0,1.5*0.618)$) rectangle ($(training)+(2,2*0.618)$);
		\filldraw[channel_selected,fill=violet_so] ($(training)+(0,1*0.618)$) rectangle ($(training)+(2,1.5*0.618)$);
		\filldraw[channel_selected,fill=green_so] ($(training)+(0,0*0.618)$) rectangle ($(training)+(2,0.5*0.618)$);
		\node at ($(training)+(1,-0.35)$) { Training Set};
		\node at ($(training)+(1,0.618*2*3/8)$) {$\cdots$};

		\coordinate (testing) at ($(big)+(4,-0.618*2)$);
		\filldraw[channel,fill=lightgray_so,fill opacity = 0.1] ($(testing)$) rectangle ($(testing)+(2,2*0.618)$);
		\filldraw[channel_selected,fill=cyan_so] ($(testing)+(0,1.5*0.618)$) rectangle ($(testing)+(2,2*0.618)$);
		\filldraw[channel_selected,fill=violet_so] ($(testing)+(0,1*0.618)$) rectangle ($(testing)+(2,1.5*0.618)$);
		\filldraw[channel_selected,fill=green_so] ($(testing)+(0,0*0.618)$) rectangle ($(testing)+(2,0.5*0.618)$);
		\node at ($(testing)+(1,-0.35)$) { Testing Set};
		\node at ($(testing)+(1,0.618*2*3/8)$) {$\cdots$};
		\draw [arrow1_thick] ($(big)+(3.1,0.618*3)$) -- ($(training)+(-0.1,0.618)$);
		\draw [arrow1_thick] ($(big)+(3.1,0)$) -- ($(testing)+(-0.1,0.618)$);
		%training testing end classifier begin
		\coordinate (classifier) at ($(big)+(7,1.5*0.618-0.618-0.25*0.618)$);
		\filldraw[channel_selected,fill=orange_so,rounded corners=2mm,fill opacity = 0.5] ($(classifier)$) rectangle ($(classifier)+(2.5,2.5*0.618)$);
		\node at ($(classifier)+(1.25,1.25*0.618)$) { Classifier};
		\draw [arrow1_thick,style=dashed] ($(training)+(2.1,0.618)$) -- ($(training)+(4.25,0.618)$)--($(classifier)+(1.25,0.618*2.5+0.1)$);
		\node at ($(training)+(3.125,0.618+0.3)$) { Train};
		\draw [arrow1_thick] ($(testing)+(2.1,0.618)$) -- ($(testing)+(4.25,0.618)$)--($(classifier)+(1.25,0-0.1)$);
		% classifier end prediction begin
		\coordinate (prediction) at ($(classifier)+(3.5,-0.25*0.618)$);
		\filldraw[channel,fill=lightgray_so,fill opacity = 0.1] ($(prediction)$) rectangle ($(prediction)+(3,3*0.618)$);
		\filldraw[channel_selected,fill=cyan_so,fill opacity = 0.7] ($(prediction)+(0,2.25*0.618)$) rectangle ($(prediction)+(3,3*0.618)$);
		\filldraw[channel_selected,fill=violet_so,fill opacity = 0.7] ($(prediction)+(0,1.5*0.618)$) rectangle ($(prediction)+(3,2.25*0.618)$);
		\filldraw[channel_selected,fill=green_so,fill opacity = 0.7] ($(prediction)+(0,0*0.618)$) rectangle ($(prediction)+(3,0.75*0.618)$);
		\node at ($(prediction)+(1.5,-0.35)$) {{Prediction}};
		\node at ($(prediction)+(1.5,0.618*3*3/8)$) {$\cdots$};
		\draw [arrow1_thick] ($(classifier)+(2.6,1.25*0.618)$) -- ($(prediction)+(-0.1,1.5*0.618)$);

		\endscope
	\end{tikzpicture}	
}
	\caption{\label{fig:flowchart} Flowchart of the proposed abundance-based multi-HSI classification method.}
\end{figure*}

In this paper, we propose an abundance-based multi-HSI classification (ABMHC) method, 
which alleviates the overfitting issues
by taking advantage of the abundance information from multiple HSIs.
To be precise, the proposed method benefits from the SU in two perspectives.
\begin{itemize}
\item \textbf{Simple network structure:} 
	The SU maps the HSI from the spectral domain to the abundance domain. 
	Different from the spectra that are usually highly mixed, the abundance features are more 
	representative in reduced dimension with less noise \cite{bioucas2012hyperspectral}. 
	By performing classification on the abundance-based features, the original classification 
	tasks are expected to be simplified, which enables the use of simple networks. 
	It is noteworthy that simple networks usually 
	have less overfitting issues \cite{hinton1993keeping}. 
\item \textbf{Enlarged training set:} 
	Transforming the HSIs into the abundance domain will eliminate 
	the data-specific information in different HSIs, {\em e.g.,} 
	the type of sensor and the spatial-spectral resolutions. 
	By considering a unified and relatively large number of endmembers 
	in SU of different HSIs, the estimated abundance features of 
	different HSIs are with the same dimension. 
	This ensures the construction of an enlarged training set, 
	that gathers the labeled data from all the HSIs in this study 
	for the subsequent classifier. 
\end{itemize}
The proposed ABMHC is generally composed of two featured procedures, 
namely 
1) SU with deep autoencoder network; 
2) CNN-based classification with extracted abundances. 
Briefly, by the dataset-wise autoencoder-based SU algorithm, the spectra from different HSIs are firstly encoded 
into abundance vectors, that are in the identical dimension.
After that, the abundance vectors from different HSIs are processed to construct  
an enlarged dataset. 
Lastly, a CNN-based classifier is trained based on the abundance patches from the enlarged dataset.  
The flowchart of the proposed ABMHC is given in \cref{fig:flowchart}.

The main contributions of the proposed ABMHC method are summarized as 
the following aspects.
\begin{itemize}
	\item We verify that performing classification over abundance representations 
	facilitates the use of simple networks, without deteriorating the performance.
	\item We verify that the classification performance is improved 
	using a unified dataset constructed from unrelated HSIs, compared with using each single HSI.
	\item We propose a method termed ABMHC that fulfills the aforementioned motivations 
	of simplifying network structure and enlarging training set.
	The proposed ABMHC is comparable to the state-of-art methods on several datasets.
\end{itemize}

The remainder of this paper is organized as follows. 
Section~\ref{sec:su} briefly presents the SU model used in this paper. 
Section~\ref{sec:autoencoder} presents 
the spectral unmixing stage with the autoencoder network, 
while Section~\ref{sec:cnn} presents 
the classification stage with CNN. 
Section \ref{sec:Experiments} reports the evaluation of 
the proposed method by ablation study and comparative experiments. 
Conclusions are drawn in Section \ref{sec:conclusion}.

\section{Notations in Spectral unmixing}\label{sec:su}

The SU consists of decomposing each observed spectrum as a mixture of 
endmembers with their proportions being abundances. 
According to different underlying mixing mechanisms, 
the SU models and associated algorithms are roughly divided into the linear 
and the nonlinear ones.
Extensive SU models and algorithms have been proposed, as reviewed 
in~\cite{bioucas2012hyperspectral,dobigeon2014nonlinear}. 
Of particular note is the recent applications of deep autoencoders for SU, 
as investigated in 
\cite{guo2015hyperspectral, wang2019nonlinear,ZhaoHyperspectral,ozkan2019endnet,su2019daen, khajehrayeni2020hyperspectral}. 
In this section, we succinctly present the SU model to be considered in this paper, 
which is proposed in \cite{wang2019nonlinear, ZhaoHyperspectral}.

Given an HSI, 
let $\bm{X} =\left [\bm{x}_1,\bm{x}_2,\ldots,\bm{x}_N\right] \in \mathbb{R}^{B\times N}$ 
be a matrix composed by $N$ observed spectra over $B$ bands, where 
$\bm{x}_i \in\mathbb{R}^B$ is the $i$-th spectrum vector, for $i=1, 2 ,... ,N$.
Assume that the HSI is known to be mixed by $R$ endmembers. 
Let $\bm{M} = \left[\bm{m}_1,\bm{m}_2,\ldots,\bm{m}_R\right] \in \mathbb{R}^{B\times R}$ 
represent  
the endmember matrix, with $\bm{m}_i$ being the spectrum of the $i$-th endmember.
The abundance vector associated with the $i$-th pixel is denoted as 
$\bm{a}_i=[a_{i,1},a_{i,2},\ldots,a_{i,R}]'$, 
its entry $a_{i,j}$ being the fractional abundance w.r.t. the $j$-th endmember, 
The linear mixing model (LMM) assumes each observed pixel to be represented as a linear combination of the endmembers, with 
\begin{equation}\label{linear_decompose}
	\bm{x}_i = \bm{M}\bm{a}_i+\bm{n},
\end{equation}
where $\bm{n} \in \mathbb{R}^B$ is the additive noise vector.
Similar to~\cite{ZhaoHyperspectral}, this paper considers a generalized SU model that 
combines the LMM and an additive nonlinear model, given by   
\begin{equation}\label{generalSU}
	\bm{x}_i = \lambda \bm{M}\bm{a}_i + (1-\lambda)\Phi(\bm{M},\bm{a}_i) + \bm{n},
\end{equation}
where $\Phi$ is a nonlinear function that characterizes the interactions between the endmembers, 
parameterized by the abundance vector,
and $\lambda$ is a hyperparameter balancing the weights of the linear and nonlinear 
parts.  
To satisfy a physical interpretation, both the abundance nonnegativity constraint (ANC)
and abundance sum-to-one constraint (ASC) are enforced to the model, 
which are
\begin{equation}\label{ANC_ASC}
	\begin{cases}
		a_{i,j}\geq 0, & j=1,2,\ldots,R; \\ 
		\sum_{j=1}^{R} a_{i,j} = 1.
	\end{cases}
\end{equation}

\section{Spectral unmixing with deep autoencoder network}\label{sec:autoencoder}
In this section, we introduce a deep autoencoder, the encoder of which 
mimics the generalized SU procedure in \cref{generalSU}, to estimate the abundance representations from the HSI. 
The proposed deep autoencoder follows the same procedure as in
\cite{wang2019nonlinear, ZhaoHyperspectral}, 
but has different network structures and implementation. 

Basically, an autoencoder is composed of two parts, namely an encoder and a decoder. 
The encoder, $\mathsf{encode}: \mathbb{R}^{B} \mapsto\mathbb{R}^{R}$, 
maps a sample from the input space to the feature space by
\begin{equation}
	\hat{\bm{a}} = \mathsf{encode}(\bm{x}).
\end{equation}
In most cases, the dimension of the input space is higher than 
that of the feature space, {\em i.e.,} $R<B$,
which indicates that the encoder compresses 
the information from input vector $\bm{x}$ to feature vector $\hat{\bm{a}}$. 
The decoder, $\mathsf{decode}: \mathbb{R}^{R} \mapsto\mathbb{R}^{B}$, 
maps the feature vector $\hat{\bm{a}}$ to an approximation $\hat{\bm{x}}$ of the original sample $\bm{x}$, 
from a low dimensional space to a high dimensional space, by
\begin{equation}
	\hat{\bm{x}} = \mathsf{decode}(\hat{\bm{a}}),
\end{equation}
where $\hat{\bm{x}}$ represents the reconstructed sample.
The whole reconstruction procedure from $\bm{x}$ to $\hat{\bm{x}}$ by the autoencoder 
is formulated as 
\begin{equation}\label{autoencoder}
	\hat{\bm{x}} = \mathsf{decode}(\mathsf{encode}(\bm{x})).
\end{equation}
When the autoencoder is good enough so that the input sample $\bm{x}$ and the 
reconstruction $\hat{\bm{x}}$ are similar under some metric, it is
inferred that the feature vector $\hat{\bm{a}}$ retrieves most of the information from $\bm{x}$.
Specifically, when the feature vector $\hat{\bm{a}}$, namely the output of the encoder, 
satisfies both the ANC and the ASC in \cref{ANC_ASC}, 
the encoder itself is interpreted 
as a blind SU procedure, and $\hat{\bm{a}}$ is taken as the abundance vector.

In this paper, both the encoder and the decoder are realized by NN. 
Let $\theta_e$ and $\theta_d$ be the learnable parameters of the encoder network and 
the decoder network, respectively. 
We use the notations $\mathsf{encode}(\cdot;\theta_e)$ and 
$\mathsf{decode}(\cdot;\theta_d,\bm{M})$ to represent the encoder and the decoder,
where the endmember matrix $\bm{M}$ is another part of learnable parameters in the decoder. 

\begin{figure}[!htb]
	\centering
{\linespread{1}
	\centering
	\tikzstyle{format}=[circle,draw,thin,fill=white]
	\tikzstyle{format_gray}=[circle,draw,thin,fill=gray]
	\tikzstyle{format_rect}=[rectangle,draw,thin,fill=white,align=center]
	\tikzstyle{arrowstyle} = [->,thick]
	\tikzstyle{network} = [rectangle, minimum width = 3cm, minimum height = 1cm, text centered, draw = black,align=center,rounded corners,fill=green_so,fill opacity=0.5,text opacity=1]
	\tikzstyle{training_batch} = [trapezium, trapezium left angle = 30, trapezium right angle = 150, minimum width = 3cm, text centered, draw = black, fill = cyan_so, fill opacity=0.3,text opacity=1,align=center]		
	\tikzstyle{class_features} = [trapezium, trapezium left angle = 30, trapezium right angle = 150, minimum width = 3cm, text centered, draw = black, fill = cyan_so, fill opacity=0.3,text opacity=1,align=center]
	\tikzstyle{pixel} = [rectangle, draw = black, fill = orange_so, fill opacity=0.5,text opacity=0,align=center]	
	\tikzstyle{pixel_red} = [rectangle, draw = black, fill = red_so, fill opacity=1,text opacity=0,align=center]	
	\tikzstyle{feature} = [rectangle, draw = black, fill = orange_so, fill opacity=0.3,text opacity=0,align=center,rounded corners]	
	\tikzstyle{feature_sfp} = [rectangle, draw = black, fill = violet_so, fill opacity=0.3,text opacity=0,align=center,rounded corners]					
	\tikzstyle{arrow1} = [thick, ->, >= stealth]
	\tikzstyle{arrow1_thick} = [thick, ->, >= stealth, line width=2pt]
	\tikzstyle{arrow2} = [thick, dashed, ->, >= stealth]
	\tikzstyle{thick_line} = [line width=0.7pt,dashed]
	\tikzstyle{channel} = [fill=white,fill opacity = 0.9]
	\tikzstyle{channel_shadow} = [fill = gray_so, fill opacity = 0.1, rounded corners]
	\tikzstyle{channel_selected} = [fill = orange_so, fill opacity = 0.8]
	\begin{tikzpicture}[auto,>=latex',  thin,  start chain=going below, every join/.style={norm}]
		\definecolor{gray_so}{RGB}{88,110,117}
		\definecolor{lightgray_so}{RGB}{207,221,221}
		\definecolor{yellow_so}{RGB}{181,137,0}
		\definecolor{cyan_so}{RGB}{42,161,152}
		\definecolor{orange_so}{RGB}{203,75,22}
		\definecolor{green_so}{RGB}{133,153,0}
		\definecolor{red_so}{RGB}{220,50,47}
		\definecolor{magenta_so}{RGB}{211,54,130}
		\definecolor{violet_so}{RGB}{108,113,196}
		\useasboundingbox (0,0) rectangle (10.7*0.8,4.7*0.8);

		\scope[transform canvas={scale=0.87}]
		
		\coordinate (zero) at (0,0);
		
		\coordinate (input) at ($(zero)+(0.7,0.75)$);
		\node at ($(input)+(0.15,-0.35)$) {{Input}};
		\node at ($(input)+(0.15,3.27)$) {$\bm{x}$};
		\filldraw[channel] ($(input)$) rectangle ($(input)+(0.25,3)$);
		\filldraw[channel_selected] ($(input)+(0,2.3)$) rectangle ($(input)+(0.25,3)$);

		\coordinate (c1) at ($(input)+(2,0)$);
		\node at($(c1)+(0.15,-0.35)$) {{Channels: $512$}};

		\foreach \x/\xtext in {19,...,0}
		{
			\pgfmathparse{Mod(\x,2)==0?1:0}
			\ifnum\pgfmathresult>0
			\filldraw[channel] ($(c1)+(\x/15,\x/15)$) rectangle ($(c1)+(0.25,2.4)+(\x/15,\x/15)$);
			\else
			\filldraw[channel,fill=lightgray_so] ($(c1)+(\x/15,\x/15)$) rectangle ($(c1)+(0.25,2.4)+(\x/15,\x/15)$);
			\fi
		}

		%words
		\node (conv1) at ($(input)!0.5!(c1)+(0.1,0.3)$){${1\textrm{D Conv}}\atop{\textrm{ReLU}}$};
		\draw [arrow1_thick] ($(input)+(0.35,0)$) -- ($(c1)+(-0.1,0)$);
		
		\filldraw[channel_selected,fill=violet_so] ($(c1)+(0,2.15)$) rectangle ($(c1)+(0.25,2.4)$);
		\filldraw[channel_selected] ($(c1)+(0,1.2)$) rectangle ($(c1)+(0.25,1.9)$);
		\draw[thick_line] ($(input)+(0.25,3)$) -- ($(c1)+(0,2.4)$);
		\draw[thick_line] ($(input)+(0.25,2.3)$)--($(c1)+(0,2.15)$);
		
		\coordinate (c2) at ($(c1)+(2,0)$);
		\node at($(c2)-(-0.25,0.35)$) {{Channels: $128$}};
		\foreach \x/\xtext in {9,...,0}
		{
			\pgfmathparse{Mod(\x,2)==0?1:0}
			\ifnum\pgfmathresult>0
			\filldraw[channel] ($(c2)+(\x/15,\x/15)$) rectangle ($(c2)+(0.25,1.8)+(\x/15,\x/15)$);
			\else
			\filldraw[channel,fill=lightgray_so] ($(c2)+(\x/15,\x/15)$) rectangle ($(c2)+(0.25,1.8)+(\x/15,\x/15)$);
			\fi
		}
		\filldraw[channel_selected,fill=violet_so] ($(c2)+(0,1.2)$) rectangle ($(c2)+(0.25,1.45)$);
		
		\node (conv2) at ($(c1)!0.5!(c2)+(0.2,0.3)$){${1\textrm{D Conv}}\atop{\textrm{ReLU}}$};
		\draw [arrow1_thick] ($(c1)+(0.35,0)$) -- ($(c2)+(-0.1,0)$);
		
		\draw[thick_line] ($(c1)+(0.25,1.2)$) -- ($(c2)+(0,1.2)$);
		\draw[thick_line] ($(c1)+(0.25,1.9)$)--($(c2)+(0,1.45)$);

		\coordinate (fc1) at ($(c2)+(2.7,0)$);
		\filldraw[channel] ($(fc1)$) rectangle ($(fc1)+(0.25,1.8)$);
		\node at($(fc1)-(-0.25,0.35)$) {{Length: $R$}};
		\draw[thick_line] ($(c2)+(0.25,0)$) -- ($(fc1)+(0,1.8)$);
		\draw[thick_line] ($(c2)+(0.25,1.8)$)--($(fc1)+(0,0)$);
		\draw[thick_line] ($(c2)+(9/15,9/15)+(0.25,0)$) -- ($(fc1)+(0,1.8)$);
		\draw[thick_line] ($(c2)+(9/15,9/15)+(0.25,1.8)$)--($(fc1)+(0,0)$);
		\node (fc1_text) at ($(c2)!0.5!(fc1)+(0.2,0.3)$){FC};
		\draw [arrow1_thick] ($(c2)+(0.35,0)$) -- ($(fc1)+(-0.1,0)$);

		\coordinate (fc2) at ($(fc1)+(2,0)$);
		\filldraw[channel] ($(fc2)$) rectangle ($(fc2)+(0.25,1.8)$);

		\node at ($(fc2)+(0.15,2.07)$) {$\hat{\bm{a}}$};
		\node at($(fc2)-(-0.25,0.35)$) {{Length: $R$}};
		\node (fc1_text) at ($(fc1)!0.5!(fc2)+(0.1,0.3)$){{Norm}};
		\draw [arrow1_thick] ($(fc1)+(0.35,0)$) -- ($(fc2)+(-0.1,0)$);
		\filldraw[channel_selected,fill=yellow_so] ($(fc2)+(0,0)$) rectangle ($(fc2)+(0.25,0.25)$);
		
		\filldraw[channel_selected,fill=violet_so] ($(fc2)+(0,1.55)$) rectangle ($(fc2)+(0.25,1.8)$);
		\filldraw[channel_selected,fill=cyan_so] ($(fc2)+(0,1.3)$) rectangle ($(fc2)+(0.25,1.55)$);
		\filldraw[channel_selected,fill=red_so] ($(fc2)+(0,1.05)$) rectangle ($(fc2)+(0.25,1.30)$);

		\endscope
	\end{tikzpicture}	
}
	\caption{\label{fig:encoder} Structure of the encoder.}
\end{figure}
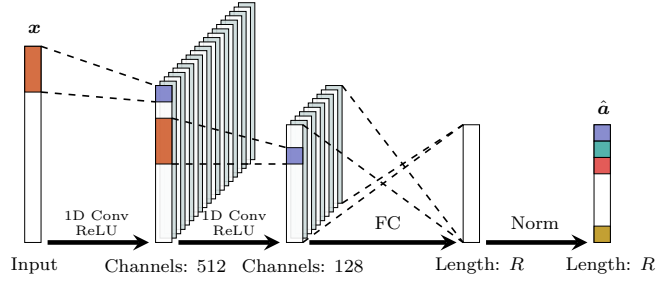

The structure of the encoder $\mathsf{encode}(\cdot;\theta_e)$ is shown in \cref{fig:encoder}. 
It has $4$ layers in total, namely two $1$D convolutional layers \cite{kiranyaz20191d}, 
one fully connected layer, 
and one normalization layer. 
The $1$D convolutional 
layer operates similarly as the plain $2$-D convolution, but the convolutional operation 
is limited to one dimension. 
In this paper, the $1$D convolutional layers are set 
with kernel size $3$, stride $1$, followed by the ReLU activations. 
The normalization layer is used to impose ANC and ASC to the
encoded abundance feature 
$\hat{\bm{a}}=\left[\hat{a}_1,\hat{a}_2,...,\hat{a}_R\right]'$ by
\begin{equation}
	{\hat{a}}_i = \frac{|\hat{a}_i|}{\sum_{j=1}^{R}|\hat{a}_j|},
\end{equation}
as suggested in \cite{wang2019nonlinear}.

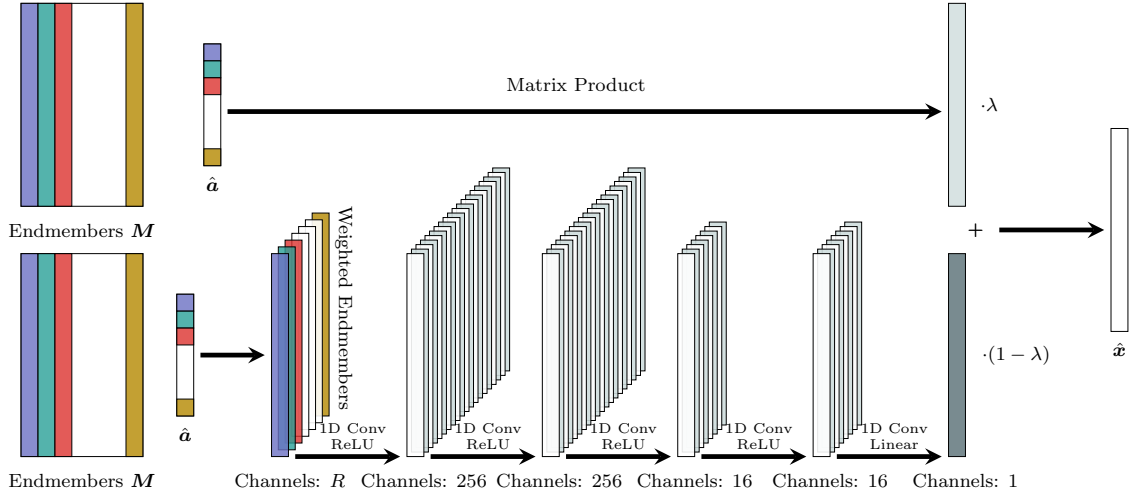
\begin{figure}[!htb]
	\centering
	{\linespread{1}
		\centering
		\tikzstyle{format}=[circle,draw,thin,fill=white]
		\tikzstyle{format_gray}=[circle,draw,thin,fill=gray]
		\tikzstyle{format_rect}=[rectangle,draw,thin,fill=white,align=center]
		\tikzstyle{arrowstyle} = [->,thick]
		\tikzstyle{network} = [rectangle, minimum width = 3cm, minimum height = 1cm, text centered, draw = black,align=center,rounded corners,fill=green_so,fill opacity=0.5,text opacity=1]
		\tikzstyle{training_batch} = [trapezium, trapezium left angle = 30, trapezium right angle = 150, minimum width = 3cm, text centered, draw = black, fill = cyan_so, fill opacity=0.3,text opacity=1,align=center]		
		\tikzstyle{class_features} = [trapezium, trapezium left angle = 30, trapezium right angle = 150, minimum width = 3cm, text centered, draw = black, fill = cyan_so, fill opacity=0.3,text opacity=1,align=center]
		\tikzstyle{pixel} = [rectangle, draw = black, fill = orange_so, fill opacity=0.5,text opacity=0,align=center]	
		\tikzstyle{pixel_red} = [rectangle, draw = black, fill = red_so, fill opacity=1,text opacity=0,align=center]	
		\tikzstyle{feature} = [rectangle, draw = black, fill = orange_so, fill opacity=0.3,text opacity=0,align=center,rounded corners]	
		\tikzstyle{feature_sfp} = [rectangle, draw = black, fill = violet_so, fill opacity=0.3,text opacity=0,align=center,rounded corners]					
		\tikzstyle{arrow1} = [thick, ->, >= stealth]
		\tikzstyle{arrow1_thick} = [thick, ->, >= stealth, line width=2pt]
		\tikzstyle{arrow2} = [thick, dashed, ->, >= stealth]
		\tikzstyle{thick_line} = [line width=0.7pt,dashed]
		\tikzstyle{channel} = [fill=white,fill opacity = 0.9]
		\tikzstyle{channel_shadow} = [fill = gray_so, fill opacity = 0.1, rounded corners]
		\tikzstyle{channel_selected} = [fill = orange_so, fill opacity = 0.8]
		\begin{tikzpicture}[auto,>=latex',  thin,  start chain=going below, every join/.style={norm}]
			\definecolor{gray_so}{RGB}{88,110,117}
			\definecolor{lightgray_so}{RGB}{207,221,221}
			\definecolor{yellow_so}{RGB}{181,137,0}
			\definecolor{cyan_so}{RGB}{42,161,152}
			\definecolor{orange_so}{RGB}{203,75,22}
			\definecolor{green_so}{RGB}{133,153,0}
			\definecolor{red_so}{RGB}{220,50,47}
			\definecolor{magenta_so}{RGB}{211,54,130}
			\definecolor{violet_so}{RGB}{108,113,196}
			\useasboundingbox (0,0) rectangle (17.7*0.87,12.7*0.618*0.87);
	
			\scope[transform canvas={scale=0.9}]
			
			\coordinate (zero) at (0,0);
			
			\coordinate (endmember) at ($(zero)+(0.7,0.75)$);

			\filldraw[channel] ($(endmember)$) rectangle ($(endmember)+(1.8,3)$);
			\filldraw[channel_selected, fill=violet_so] ($(endmember)$) rectangle ($(endmember)+(0.25,3)$);
			\filldraw[channel_selected, fill=cyan_so] ($(endmember)+(0.25,0)$) rectangle ($(endmember)+(0.5,3)$);
			\filldraw[channel_selected, fill=red_so] ($(endmember)+(0.5,0)$) rectangle ($(endmember)+(0.75,3)$);
			\filldraw[channel_selected, fill=yellow_so] ($(endmember)+(1.55,0)$) rectangle ($(endmember)+(1.8,3)$);
			\node at ($(endmember)+(0.9,-0.35)$) {Endmembers $\bm{M}$};

			\coordinate (abundance) at ($(endmember)+(2.3,0.6)$);
			\node at ($(abundance)+(0.127,-0.27)$) {$\hat{\bm{a}}$};
			\filldraw[channel] ($(abundance)$) rectangle ($(abundance)+(0.25,1.8)$);
			\filldraw[channel_selected,fill=violet_so] ($(abundance)+(0,1.55)$) rectangle ($(abundance)+(0.25,1.8)$);
			\filldraw[channel_selected,fill=cyan_so] ($(abundance)+(0,1.3)$) rectangle ($(abundance)+(0.25,1.55)$);
			\filldraw[channel_selected,fill=red_so] ($(abundance)+(0,1.05)$) rectangle ($(abundance)+(0.25,1.30)$);
			\filldraw[channel_selected,fill=yellow_so] ($(abundance)+(0,0)$) rectangle ($(abundance)+(0.25,0.25)$);

			\coordinate (c1) at ($(endmember)+(3.7,0)$);
			\node at ($(c1)+(0.27,-0.35)$) {{Channels: $R$}};
			\node [rotate=-90] at ($(c1)+(1.07,2.2)$) {Weighted Endmembers};
			\filldraw[channel_selected,fill=yellow_so] ($(c1)+(6/10,6/10)$) rectangle ($(c1)+(0.25,3)+(6/10,6/10)$);
			\filldraw[channel] ($(c1)+(5/10,5/10)$) rectangle ($(c1)+(0.25,3)+(5/10,5/10)$);
			\filldraw[channel] ($(c1)+(4/10,4/10)$) rectangle ($(c1)+(0.25,3)+(4/10,4/10)$);
			\filldraw[channel] ($(c1)+(3/10,3/10)$) rectangle ($(c1)+(0.25,3)+(3/10,3/10)$);
			\filldraw[channel_selected,fill=red_so] ($(c1)+(2/10,2/10)$) rectangle ($(c1)+(0.25,3)+(2/10,2/10)$);
			\filldraw[channel_selected,fill=cyan_so] ($(c1)+(1/10,1/10)$) rectangle ($(c1)+(0.25,3)+(1/10,1/10)$);
			\filldraw[channel_selected,fill=violet_so] ($(c1)+(0/10,0/10)$) rectangle ($(c1)+(0.25,3)+(0/10,0/10)$);
			\draw [arrow1_thick] ($(abundance)+(0.35,0.9)$) -- ($(c1)+(-0.1,1.5)$);

			\coordinate (c2) at ($(c1)+(2,0)$);
			\node at($(c2)-(-0.25,0.35)$) {{Channels: $256$}};
			\foreach \x/\xtext in {19,...,0}
			{
				\pgfmathparse{Mod(\x,2)==0?1:0}
				\ifnum\pgfmathresult>0
				\filldraw[channel] ($(c2)+(\x/15,\x/15)$) rectangle ($(c2)+(0.25,3)+(\x/15,\x/15)$);
				\else
				\filldraw[channel,fill=lightgray_so] ($(c2)+(\x/15,\x/15)$) rectangle ($(c2)+(0.25,3)+(\x/15,\x/15)$);
				\fi
			}
			
			\node (conv2) at ($(c1)!0.5!(c2)+(0.2,0.3)$){${1\textrm{D Conv}}\atop{\textrm{ReLU}}$};
			\draw [arrow1_thick] ($(c1)+(0.35,0)$) -- ($(c2)+(-0.1,0)$);

			\coordinate (c3) at ($(c2)+(2,0)$);
			\node at($(c3)-(-0.25,0.35)$) {{Channels: $256$}};
			\foreach \x/\xtext in {19,...,0}
			{
				\pgfmathparse{Mod(\x,2)==0?1:0}
				\ifnum\pgfmathresult>0
				\filldraw[channel] ($(c3)+(\x/15,\x/15)$) rectangle ($(c3)+(0.25,3)+(\x/15,\x/15)$);
				\else
				\filldraw[channel,fill=lightgray_so] ($(c3)+(\x/15,\x/15)$) rectangle ($(c3)+(0.25,3)+(\x/15,\x/15)$);
				\fi
			}
			
			\node (conv2) at ($(c2)!0.5!(c3)+(0.2,0.3)$){${1\textrm{D Conv}}\atop{\textrm{ReLU}}$};
			\draw [arrow1_thick] ($(c2)+(0.35,0)$) -- ($(c3)+(-0.1,0)$);

			\coordinate (c4) at ($(c3)+(2,0)$);
			\node at($(c4)-(-0.25,0.35)$) {{Channels: $16$}};
			\foreach \x/\xtext in {7,...,0}
			{
				\pgfmathparse{Mod(\x,2)==0?1:0}
				\ifnum\pgfmathresult>0
				\filldraw[channel] ($(c4)+(\x/15,\x/15)$) rectangle ($(c4)+(0.25,3)+(\x/15,\x/15)$);
				\else
				\filldraw[channel,fill=lightgray_so] ($(c4)+(\x/15,\x/15)$) rectangle ($(c4)+(0.25,3)+(\x/15,\x/15)$);
				\fi
			}
			\node (conv2) at ($(c3)!0.5!(c4)+(0.2,0.3)$){${1\textrm{D Conv}}\atop{\textrm{ReLU}}$};
			\draw [arrow1_thick] ($(c3)+(0.35,0)$) -- ($(c4)+(-0.1,0)$);

			\coordinate (c5) at ($(c4)+(2,0)$);
			\node at($(c5)-(-0.25,0.35)$) {{Channels: $16$}};
			\foreach \x/\xtext in {7,...,0}
			{
				\pgfmathparse{Mod(\x,2)==0?1:0}
				\ifnum\pgfmathresult>0
				\filldraw[channel] ($(c5)+(\x/15,\x/15)$) rectangle ($(c5)+(0.25,3)+(\x/15,\x/15)$);
				\else
				\filldraw[channel,fill=lightgray_so] ($(c5)+(\x/15,\x/15)$) rectangle ($(c5)+(0.25,3)+(\x/15,\x/15)$);
				\fi
			}
			\node (conv2) at ($(c4)!0.5!(c5)+(0.2,0.3)$){${1\textrm{D Conv}}\atop{\textrm{ReLU}}$};
			\draw [arrow1_thick] ($(c4)+(0.35,0)$) -- ($(c5)+(-0.1,0)$);

			\coordinate (c6) at ($(c5)+(2,0)$);
			\node at($(c6)-(-0.25,0.35)$) {{Channels: $1$}};
			\node at ($(c6)+(1,1.5)$) {$\cdot(1-\lambda)$};
			\filldraw[channel_selected,fill=gray_so] ($(c6)$) rectangle ($(c6)+(0.25,3)$);
			\node (conv2) at ($(c5)!0.5!(c6)+(0.2,0.3)$){${1\textrm{D Conv}}\atop{\textrm{Linear}}$};
			\draw [arrow1_thick] ($(c5)+(0.35,0)$) -- ($(c6)+(-0.1,0)$);

			%% upper procudeure
			\coordinate (endmember_u) at ($(endmember)+(0,3.7)$);

			\filldraw[channel] ($(endmember_u)$) rectangle ($(endmember_u)+(1.8,3)$);
			\filldraw[channel_selected, fill=violet_so] ($(endmember_u)$) rectangle ($(endmember_u)+(0.25,3)$);
			\filldraw[channel_selected, fill=cyan_so] ($(endmember_u)+(0.25,0)$) rectangle ($(endmember_u)+(0.5,3)$);
			\filldraw[channel_selected, fill=red_so] ($(endmember_u)+(0.5,0)$) rectangle ($(endmember_u)+(0.75,3)$);
			\filldraw[channel_selected, fill=yellow_so] ($(endmember_u)+(1.55,0)$) rectangle ($(endmember_u)+(1.8,3)$);
			\node at ($(endmember_u)+(0.9,-0.35)$) {Endmembers $\bm{M}$};

			\coordinate (abundance_u) at ($(endmember_u)+(2.7,0.6)$);
			\node at ($(abundance_u)+(0.127,-0.27)$) {$\hat{\bm{a}}$};
			% \node at ($(abundance_u)+(-0.45,0.8)$) {$\bullet$};
			\filldraw[channel] ($(abundance_u)$) rectangle ($(abundance_u)+(0.25,1.8)$);
			\filldraw[channel_selected,fill=violet_so] ($(abundance_u)+(0,1.55)$) rectangle ($(abundance_u)+(0.25,1.8)$);
			\filldraw[channel_selected,fill=cyan_so] ($(abundance_u)+(0,1.3)$) rectangle ($(abundance_u)+(0.25,1.55)$);
			\filldraw[channel_selected,fill=red_so] ($(abundance_u)+(0,1.05)$) rectangle ($(abundance_u)+(0.25,1.30)$);
			\filldraw[channel_selected,fill=yellow_so] ($(abundance_u)+(0,0)$) rectangle ($(abundance_u)+(0.25,0.25)$);

			\coordinate (c6_u) at ($(c6)+(endmember_u)-(endmember)$);
			\filldraw[channel_selected,fill=lightgray_so] ($(c6_u)$) rectangle ($(c6_u)+(0.25,3)$);
			\node at ($(c6_u)+(0.6,1.5)$) {$\cdot\lambda$};
			\node (plussign) at ($(c6_u)+(0.4,-0.35)$) {$+$};
			\node at ($(c6_u)+(0.4,-0.35)$) {$+$};
			\draw [arrow1_thick] ($(abundance_u)+(0.35,0.8)$) -- ($(c6_u)+(-0.1,1.4)$);
			\node at ($(abundance_u)!0.5!(c6_u)+(0,1.5)$) {Matrix Product};

			\coordinate (reconstruction) at ($(plussign)+(2,-1.5)$);
			\node at ($(reconstruction)+(0.127,-0.27)$) {$\hat{\bm{x}}$};
			\filldraw[channel] ($(reconstruction)$) rectangle ($(reconstruction)+(0.25,3)$);
			\draw [arrow1_thick] ($(plussign)+(0.35,0)$) -- ($(reconstruction)+(-0.1,1.5)$);
			\endscope
		\end{tikzpicture}
	}
		\caption{\label{fig:decoder} Structure of the decoder.}
\end{figure}

The structure of the decoder $\mathsf{decode}(\cdot;\theta_d,\bm{M})$ is more complicated. 
It is designed to consist of two parts, 
that correspond to the linear and nonlinear mixing models, 
in accordance with the latent mixing mechanism in~\cref{generalSU}. 
As illustrated in~\cref{fig:decoder}, 
the upper part corresponds to the LMM, expressed by $\bm{M}\bm{a}$, 
while the lower part represents the nonlinear mixing model given by $\Phi(\bm{M},\bm{a})$. 
For the nonlinear part, we first multiply each endmember by its fractional abundance, 
thus generating a set of weighted endmembers, 
given by $[\hat{a}_1\bm{m}_1,\hat{a}_2\bm{m}_2,\ldots,\hat{a}_R\bm{m}_R]$.
Later, the weighted endmembers flow through five $1$D convolutional 
layers with different numbers of output channels and end up 
with a vector of length $B$. 
All the five $1$D convolutional layers are set to have kernel size $1$ and 
stride $1$. This setting ensures that the effect of the $1$D convolutional layer 
could be interpreted by the interaction between the channels of the input data. 
In view of this, the nonlinear part of the decoder simulates the interactions 
between the endmember signatures. 
Each of the $1$D convolutional layers is 
followed by the ReLU activation, except for the last one. 
In the end, the reconstructed spectrum is estimated by 
the weighted sum of the linear and nonlinear estimations.

In the proposed method, for each HSI scene, we propose to train a specific pair of 
encoder and decoder to represent the blind SU procedure and the
spectrum reconstruction, respectively. 
The encoder $\mathsf{encode}(\bm{x};\theta_e)$ maps the spectrum $\bm{x}$ to 
its abundance vector $\hat{\bm{a}} = \mathsf{encode}(\bm{x};\theta_e)$, where $\theta_e$ represents 
the network parameters with the implicit endmember matrix $\bm{M}$. 
The decoder $\mathsf{decode}(\hat{\bm{a}};\theta_d,\bm{M})$ takes the abundance vector $\hat{\bm{a}}$ 
as inputs and outputs the reconstructed spectrum $\hat{\bm{x}}=\mathsf{decode}(\hat{\bm{a}};\theta_d,\bm{M})$, and contains two 
parts of parameters, namely the parameters of the neural network $\theta_d$ and the explicit endmember matrix $\bm{M}$.
To learn the parameters $\theta_e,\theta_d$ and endmember matrix $\bm{M}$, we jointly optimize the 
encoder and decoder
\begin{equation}
	\hat{\bm{x}} = \mathsf{decode}(\mathsf{encode}(\bm{x};\theta_e);\theta_d,\bm{M}), 
\end{equation}
by minimizing the reconstruction error 
\begin{equation}\label{eq:RE}
	\mathcal{L}_r = \frac{\sum_{i=1}^{N} ||\bm{x}_i-\hat{\bm{x}}_i ||_2^2}{N},
\end{equation}
where $\bm{x}_i$ is the spectrum from the HSI scene and $\hat{\bm{x}}_i$ is the 
reconstructed spectrum corresponding to $\bm{x}_i$.
To optimize \cref{eq:RE}, different gradient-based algorithms can be applied 
\cite{bottou2010large,kingma2015adam,dubey2019diffgrad}.
The endmember matrix $\bm{M}$ is initialized by the vertex component 
analysis (VCA) \cite{nascimento2005vertex},
and the paramerters $\theta_e$ and $\theta_d$ are initialized by the uniform distribution, as in \cite{wang2019nonlinear}. 
The optimized parameters are denoted as 
$\hat{\theta}_e,\hat{\theta}_d$ and $\hat{\bm{M}}$, and
hereafter we use 
$\mathsf{encode}(\cdot)$ and $\mathsf{decode}(\cdot)$ 
to denote the learned models
$\mathsf{encode}(\cdot;\hat{\theta}_e)$ and 
$\mathsf{decode}(\cdot;\hat{\theta}_d,\hat{\bm{M}})$, respectively, 
unless otherwise stated.

Before classification, we utilize every learned encoder to map its corresponding HSI 
to the abundance representation by
\begin{equation}
    \hat{\bm{A}} = \mathsf{encode}(\bm{X}).
\end{equation}
Accounting for the subsequent classification tasks on multiple HSIs, 
we have the following settings on the autoencoder-based SU procedure.
The number of endmembers is set to be $R=16$ for all the autoencoders, 
a value larger than the real numbers of endmembers for most HSIs. 
The redundancy of endmembers ensures that the abundances could have the capacity 
to retrieve most of the information from spectra.
The hyperparameter $\lambda$ is set to be $0.5$, 
following \cite{wang2019nonlinear, ZhaoHyperspectral}.
The HSIs are normalized to the range $[0,1]$ before fed to the 
autoencoder.
As the research focus of this work is not the SU model and method, 
the analysis of the effects of these hyperparameters is omitted. 
Readers may refer to \cite{wang2019nonlinear} for a detailed hyperparameter analysis 
of a similar SU procedure. 
The autoencoder is optimized by the Adam algorithm \cite{kingma2015adam}. 
The structures of networks are designed and realized with AutoKeras \cite{jin2019auto}
and Tensorflow \cite{abadi2016tensorflow}.

\section{Multi-HSI classification with convolutional neural network}\label{sec:cnn}

In this section, we use a simple CNN model based on both the spatial information and 
the abundance representations to jointly classify multiple HSIs. 
To alleviate the overfitting issue, merging different HSIs into one big dataset 
is one of the most fundamental motivations of this paper. 
Different from existing CNN models, which process the raw data from single HSI, 
the proposed algorithm is capable to process the 
abundance data from multiple HSIs simultaneously. 

\subsection{Preparation of training and testing data}\label{PrepareTraining}

To construct a big dataset from different HSIs, 
we propose the following processings of the autoencoder-extracted abundance representations. 
Given $K$ HSIs to be classified, namely $\{\bm{X}^{k}|\,k=1,2,\ldots,K\}$, 
assume that $\bm{X}^{k}$ contains $C^{k}$ labeled classes. 
By training an autoencoder $\mathsf{decode}^k(\mathsf{encode}^k(\cdot))$, 
which is described in \cref{sec:autoencoder}, for every HSI separately, 
we obtain the following abundance representations
\begin{equation}
	\hat{\bm{A}}^{k} = \mathsf{encode}^{k}(\bm{X}^{k}),\quad k=1,2,\ldots,K. 
\end{equation}
To take advantage of both spatial context and 
abundance information for improving the classification performance, 
each abundance representation $\hat{\bm{A}}^{k}$ is 
firstly divided into the labeled abundance patches 
\begin{equation}
	\{(\hat{\mathfrak{a}}^k_i,y^k_i)|\, i=1,2,\ldots,N^k\}, 
\end{equation}
where $\hat{\mathfrak{a}}^k_i$ represents an abundance patch from $\hat{\bm{A}}^{k}$, 
and corresponds to a pixel patch in the original image $\bm{X}^{k}$. 
The label $y^k_i$ is selected as the label of the pixel patch center, 
and ranges from $0$ to $C^k-1$.

Assume we have following $K$ sets of labeled abundance patches generated from $K$ HSIs,
\begin{eqnarray*}
	\{(\hat{\mathfrak{a}}^1_i,y^1_i)|\, i=1,2,\ldots,N^1\}, && y^1_i\in \{0,1,\ldots,C^1-1\};\\
	\{(\hat{\mathfrak{a}}^2_i,y^2_i)|\, i=1,2,\ldots,N^2\}, && y^2_i\in \{0,1,\ldots,C^2-1\};\\
	&\vdots&\\
	\{(\hat{\mathfrak{a}}^K_i,y^K_i)|\, i=1,2,\ldots,N^K\}, && y^K_i\in \{0,1,\ldots,C^K-1\}. 
\end{eqnarray*}
A big merged dataset $S$ is constructed by collecting all sets of labeled abundance patches, 
with labels rearranged to avoid overlap. 
To be precise, the labeled abundance patch $(\hat{\mathfrak{a}}^k_i,y^k_i)$ is 
relabeled to 
\begin{equation}
	\left(\hat{\mathfrak{a}}^k_i,y^k_i+\sum_{j=1}^{k-1}C^j\right)
\end{equation}
before collected into $S$. 
By doing so, in the merged dataset, each HSI 
occupies a specific interval of integers as class labels, without mutual 
overlaps of labels with other HSIs. 
In summary, the big dataset $S$ assembles all the samples from 
$\hat{\bm{A}}^{k}, k=1,2,\ldots,K$, with $C=\sum_{k=1}^{K}C^k$ 
being the total number of classes.

\subsection{Classification with CNN}

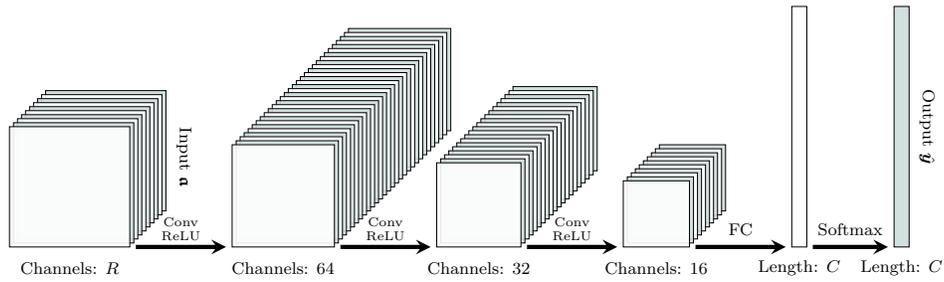
\begin{figure}[!htb]
	\centering
{\linespread{1}
	\centering
	\tikzstyle{format}=[circle,draw,thin,fill=white]
	\tikzstyle{format_gray}=[circle,draw,thin,fill=gray]
	\tikzstyle{format_rect}=[rectangle,draw,thin,fill=white,align=center]
	\tikzstyle{arrowstyle} = [->,thick]
	\tikzstyle{network} = [rectangle, minimum width = 3cm, minimum height = 1cm, text centered, draw = black,align=center,rounded corners,fill=green_so,fill opacity=0.5,text opacity=1]
	\tikzstyle{training_batch} = [trapezium, trapezium left angle = 30, trapezium right angle = 150, minimum width = 3cm, text centered, draw = black, fill = cyan_so, fill opacity=0.3,text opacity=1,align=center]		
	\tikzstyle{class_features} = [trapezium, trapezium left angle = 30, trapezium right angle = 150, minimum width = 3cm, text centered, draw = black, fill = cyan_so, fill opacity=0.3,text opacity=1,align=center]
	\tikzstyle{pixel} = [rectangle, draw = black, fill = orange_so, fill opacity=0.5,text opacity=0,align=center]	
	\tikzstyle{pixel_red} = [rectangle, draw = black, fill = red_so, fill opacity=1,text opacity=0,align=center]	
	\tikzstyle{feature} = [rectangle, draw = black, fill = orange_so, fill opacity=0.3,text opacity=0,align=center,rounded corners]	
	\tikzstyle{feature_sfp} = [rectangle, draw = black, fill = violet_so, fill opacity=0.3,text opacity=0,align=center,rounded corners]					
	\tikzstyle{arrow1} = [thick, ->, >= stealth]
	\tikzstyle{arrow1_thick} = [thick, ->, >= stealth, line width=2pt]
	\tikzstyle{arrow2} = [thick, dashed, ->, >= stealth]
	\tikzstyle{thick_line} = [line width=0.7pt,dashed]
	\tikzstyle{channel} = [fill=white,fill opacity = 0.9]
	\tikzstyle{channel_shadow} = [fill = gray_so, fill opacity = 0.1, rounded corners]
	\tikzstyle{channel_selected} = [fill = orange_so, fill opacity = 0.8]
	\begin{tikzpicture}[auto,>=latex',  thin,  start chain=going below, every join/.style={norm}]
		\definecolor{gray_so}{RGB}{88,110,117}
		\definecolor{lightgray_so}{RGB}{207,221,221}
		\definecolor{yellow_so}{RGB}{181,137,0}
		\definecolor{cyan_so}{RGB}{42,161,152}
		\definecolor{orange_so}{RGB}{203,75,22}
		\definecolor{green_so}{RGB}{133,153,0}
		\definecolor{red_so}{RGB}{220,50,47}
		\definecolor{magenta_so}{RGB}{211,54,130}
		\definecolor{violet_so}{RGB}{108,113,196}
		\useasboundingbox (0,0) rectangle (16.7*0.8,5.*0.8);

		\scope[transform canvas={scale=0.8}]
		
		\coordinate (zero) at (0,0);
		
		\coordinate (input) at ($(zero)+(0.7,0.75)$);
		\node [rotate=-90] at ($(input)+(2.9,1.5)$) {Input $\bm{\mathfrak{a}}$};
		\node at ($(input)+(1,-0.35)$) {{Channels: $R$}};
		\foreach \x/\xtext in {9,...,0}
		{
			\pgfmathparse{Mod(\x,2)==0?1:0}
			\ifnum\pgfmathresult>0
			\filldraw[channel] ($(input)+(\x/15,\x/15)$) rectangle ($(input)+(2,2)+(\x/15,\x/15)$);
			\else
			\filldraw[channel,fill=lightgray_so] ($(input)+(\x/15,\x/15)$) rectangle ($(input)+(2,2)+(\x/15,\x/15)$);
			\fi
		}

		\coordinate (c1) at ($(input)+(2,0)+(1.7,0)$);
		\node at($(c1)+(0.85,-0.35)$) {{Channels: $64$}};

		\foreach \x/\xtext in {29,...,0}
		{
			\pgfmathparse{Mod(\x,2)==0?1:0}
			\ifnum\pgfmathresult>0
			\filldraw[channel] ($(c1)+(\x/15,\x/15)$) rectangle ($(c1)+(1.7,1.7)+(\x/15,\x/15)$);
			\else
			\filldraw[channel,fill=lightgray_so] ($(c1)+(\x/15,\x/15)$) rectangle ($(c1)+(1.7,1.7)+(\x/15,\x/15)$);
			\fi
		}

		% \node (conv1) at ($(input)!0.5!(c1)+(0.1,0.3)$){${1\textrm{D Conv}}\atop{\textrm{ReLU}}$};
		\coordinate (arrow1_s) at ($(input)+(2,0)+(0.1,0)$);
		\coordinate (arrow1_e) at ($(c1)+(-0.1,0)$);
		\draw [arrow1_thick] ($(arrow1_s)$) -- ($(arrow1_e)$);
		\node at ($(arrow1_s)!0.5!(arrow1_e) + (0,0.3)$) {${\textrm{Conv}}\atop{\textrm{ReLU}}$};

		\coordinate (c2) at ($(c1)+(1.7,0)+(1.7,0)$);
		\node at($(c2)+(0.7,-0.35)$) {{Channels: $32$}};
		\foreach \x/\xtext in {19,...,0}
		{
			\pgfmathparse{Mod(\x,2)==0?1:0}
			\ifnum\pgfmathresult>0
			\filldraw[channel] ($(c2)+(\x/15,\x/15)$) rectangle ($(c2)+(1.4,1.4)+(\x/15,\x/15)$);
			\else
			\filldraw[channel,fill=lightgray_so] ($(c2)+(\x/15,\x/15)$) rectangle ($(c2)+(1.4,1.4)+(\x/15,\x/15)$);
			\fi
		}
		
		\coordinate (arrow2_s) at ($(c1)+(1.7,0)+(0.1,0)$);
		\coordinate (arrow2_e) at ($(c2)+(-0.1,0)$);
		\draw [arrow1_thick] ($(arrow2_s)$) -- ($(arrow2_e)$);
		\node at ($(arrow2_s)!0.5!(arrow2_e) + (0,0.3)$) {${\textrm{Conv}}\atop{\textrm{ReLU}}$};

		\coordinate (c3) at ($(c2)+(1.4,0)+(1.7,0)$);
		\node at($(c3)+(0.55,-0.35)$) {{Channels: $16$}};
		\foreach \x/\xtext in {9,...,0}
		{
			\pgfmathparse{Mod(\x,2)==0?1:0}
			\ifnum\pgfmathresult>0
			\filldraw[channel] ($(c3)+(\x/15,\x/15)$) rectangle ($(c3)+(1.1,1.1)+(\x/15,\x/15)$);
			\else
			\filldraw[channel,fill=lightgray_so] ($(c3)+(\x/15,\x/15)$) rectangle ($(c3)+(1.1,1.1)+(\x/15,\x/15)$);
			\fi
		}
		\coordinate (arrow3_s) at ($(c2)+(1.4,0)+(0.1,0)$);
		\coordinate (arrow3_e) at ($(c3)+(-0.1,0)$);
		\draw [arrow1_thick] ($(arrow3_s)$) -- ($(arrow3_e)$);
		\node at ($(arrow3_s)!0.5!(arrow3_e) + (0,0.3)$) {${\textrm{Conv}}\atop{\textrm{ReLU}}$};

		\coordinate (fc1) at ($(c3)+(1.1,0)+(1.7,0)$);
		\node at ($(fc1)+(0.125,-0.35)$) {{Length: $C$}};
		\filldraw[channel] ($(fc1)$) rectangle ($(fc1)+(0.25,4)$);
		\coordinate (arrow4_s) at ($(c3)+(1.1,0)+(0.1,0)$);
		\coordinate (arrow4_e) at ($(fc1)+(-0.1,0)$);
		\draw [arrow1_thick] ($(arrow4_s)$) -- ($(arrow4_e)$);
		\node at ($(arrow4_s)!0.5!(arrow4_e) + (0,0.3)$) {FC};

		\coordinate (softmax) at ($(fc1)+(1.7,0)$);
		\node at ($(softmax)+(0.125,-0.35)$) {{Length: $C$}};
		\node [rotate=-90] at ($(softmax)+(0.55,2)$) {Output $\hat{\bm{y}}$};
		\filldraw[channel,fill=lightgray_so] ($(softmax)$) rectangle ($(softmax)+(0.25,4)$);
		\coordinate (arrow5_s) at ($(fc1)+(0.25,0)+(0.1,0)$);
		\coordinate (arrow5_e) at ($(softmax)+(-0.1,0)$);
		\draw [arrow1_thick] ($(arrow5_s)$) -- ($(arrow5_e)$);
		\node at ($(arrow5_s)!0.5!(arrow5_e) + (0,0.3)$) {Softmax};
		\endscope
	\end{tikzpicture}	
}
	\caption{\label{fig:classifier} Structure of the classifier.}
\end{figure}

Given a dataset $S$ and a sample $(\bm{\mathfrak{a}},y)\in S$ to classify, 
the classification task is interpreted as finding a function 
$\mathsf{classify}(\cdot)$ that maps the abundance patch $\bm{\mathfrak{a}}$ 
to the correct label $y$. 
This task is realized by a simple neural network 
that consists of most of the well-known CNN layers. 
As illustrated in \cref{fig:classifier}, the network 
takes abundance data, which has $R$ channels, as input. 
Later, the data flows through three CNN layers, 
each followed by a ReLU activation. 
These CNN layers are set with kernel size $3\times 3$ and stride $1$.  
In the last two steps, a fully connected layer 
maps the data into a vector with length $C$, namely the total number of classes in $S$; 
and a softmax layer finally transforms the vector into the output with one-hot style. 
The softmax function produces the predicted probability distribution of sample $\bm{\mathfrak{a}}$
as follows 
\begin{equation}
	\mathsf{softmax}(\bm{x})_i = \frac{\exp(x_i)}{\sum_{j=0}^{C-1} \exp(x_j)},\quad {i=0,1,\ldots,C-1}.
\end{equation}

Separate the merged dataset $S$ into the training set $S_{\mathrm{tr}}$ and 
the testing set $S_{\mathrm{te}}$. 
Let $\theta$ be the parameters of the proposed CNN, and 
$\hat{y}$ be the prediction on $\bm{\mathfrak{a}}$. 
In the training stage, the parameters $\theta$ are optimized by minimizing the following
cross-entropy loss
\begin{equation}\label{eq:loss}
	\mathcal{L}_s = \sum_{(\bm{\mathfrak{a}},y)\in S_{\mathrm{tr}}}
	\sum_{i=0}^{C-1} -y_i\log{\hat{y}_i},
\end{equation}
over the training set, 
where $\bm{y} = [y_0,y_1,\ldots,y_{C-1}]'$ and 
$\bm{\hat{y}} = [\hat{y}_0,\hat{y}_1,\ldots,\hat{y}_{C-1}]'$ are the 
one-hot encodes of $y$ and $\hat{y}$, respectively. 
After minimizing \cref{eq:loss} by gradient-based optimization,
the optimized parameters $\hat{\theta}$ are obtained. 
Hereafter the trained classifier $\mathsf{classify}(\cdot;\hat{\theta})$ is abbreviated by
$\mathsf{classify}(\cdot)$.

In the testing stage, when a testing sample $\bm{\mathfrak{a}}$ is fed into the trained classifier 
$\mathsf{classify}(\cdot)$, the one-hot style prediction is generated as
\begin{eqnarray}
	\bm{\hat{y}} &=&  \mathsf{classify}(\bm{\mathfrak{a}}) \\
	&=& [\hat{y}_0,\hat{y}_1,\ldots,\hat{y}_{C-1}]'.
\end{eqnarray}
In most of the existing classification models, the final prediction $\hat{y}$ is calculated by 
\begin{equation}
	\hat{y} = \argmax_i \{\hat{y}'_i|\, i=0,1,\ldots,C-1\}.
\end{equation}
However, in this paper, as we know in advance from which HSI the sample comes, 
an elaborated strategy is applied to further improve 
the accuracy of testing.
Assume the testing sample $\bm{\mathfrak{a}}$ in known to be from the 
abundance representation $\bm{A}^k$ of the $k$-th HSI. 
The predicted label of $\bm{\mathfrak{a}}$ is calculated by performing 
$\argmax$ function merely on the fragment of $\bm{\hat{y}}$ corresponding to $\bm{A}^k$, by
\begin{equation}
	\hat{y} = \argmax_i \left\{\hat{y}'_i\left|\, i=\sum_{j=1}^{k-1}C^j,\sum_{j=1}^{k-1}C^j+1,\ldots,\sum_{j=1}^{k}C^j-1\right.\right\}.
\end{equation}

\section{Experiments}\label{sec:Experiments}
In this section, we perform a series of experiments including 
the ablation study and comparative study with several state-of-the-art methods 
on four public HSI datasets, to verify 
the effectiveness of the innovative ideas and 
to demonstrate the performance of 
the proposed method. 
\subsection{Datasets}
In this paper, experiments are performed on four public HSI datasets, {\em i.e.},
the Paiva University scene\footnote{The datasets 
are available online: 
\url{http://www.ehu.eus/ccwintco/index.php?title=Hyperspectral_Remote_Sensing_Scenes}\label{footnote}}, 
the Pavia Centre scene\textsuperscript{\ref{footnote}}, 
the Salinas scene\textsuperscript{\ref{footnote}},
and the Houston2018 scene\footnote{This data is from 2018 IEEE GRSS Data Fusion Contest, where only the training set is used in this paper. 
The dataset is available online: \url{http://www.grss-ieee.org/community/technical-committees/data-fusion}}
(grss\_dfc\_2018) \cite{xu2019advanced}. 

\begin{figure}[!htb]
	\centering
\graphicspath{{Figures/}}
	\emph{\includegraphics[trim = 25mm 30mm 31mm 25mm, clip,width=.2\textwidth]  {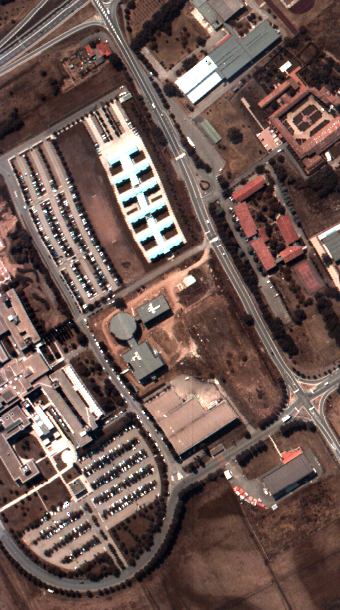}
		\includegraphics[trim = 25mm 30mm 31mm 25mm,clip,width=.2\textwidth]  {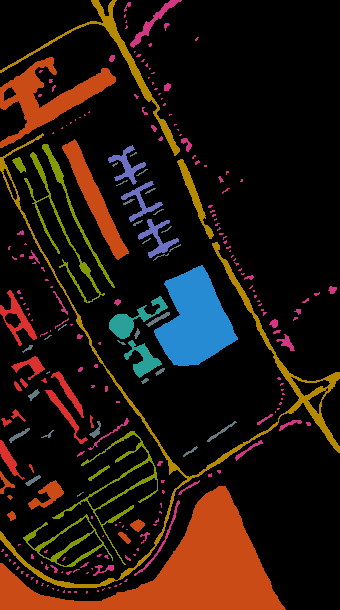}
	}\includegraphics[trim = 0mm 0mm 0mm 0mm,clip,width=.21\textwidth]  {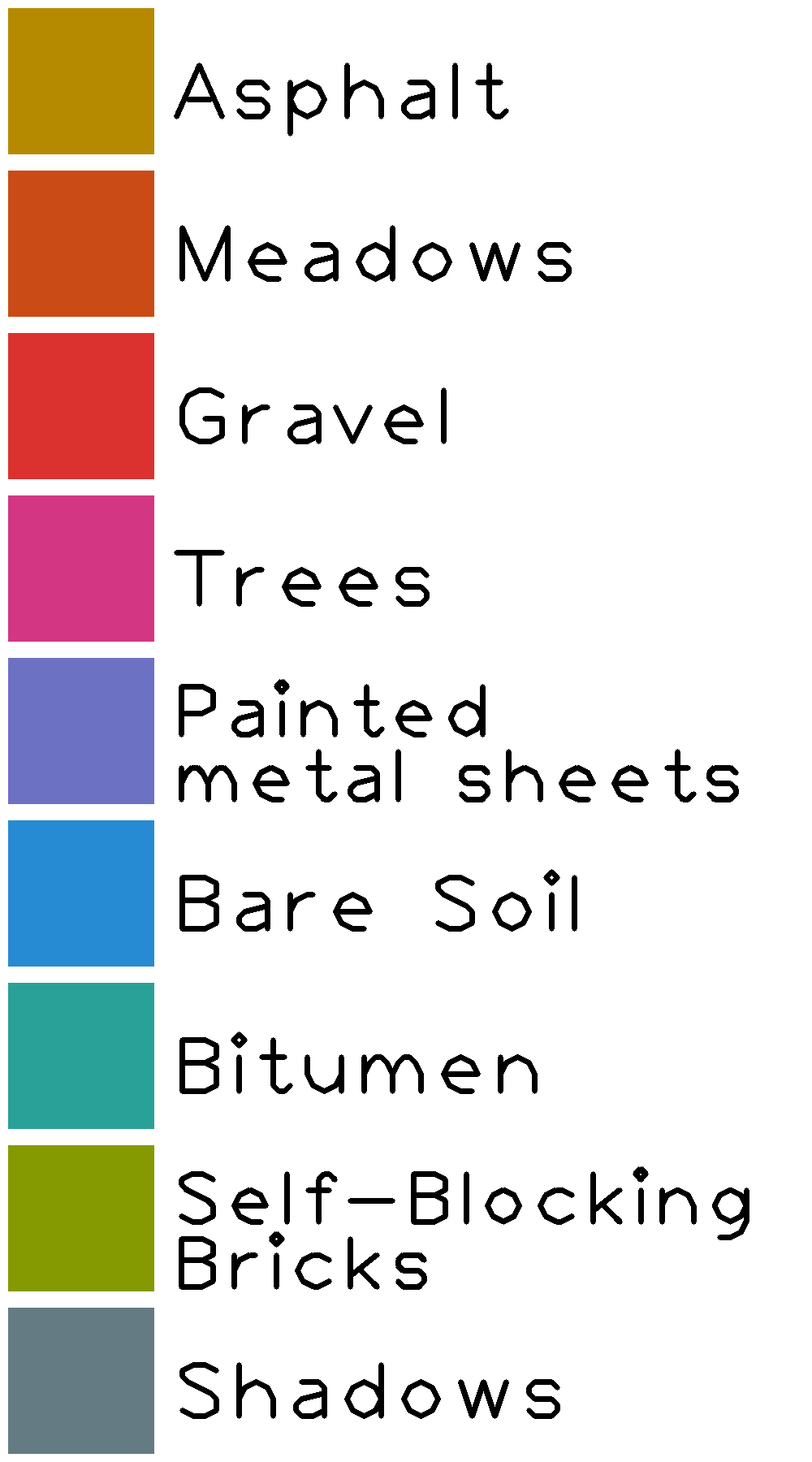}
	\caption{\label{paviau_fc_gt} The false color composite (band $10,20,40$) and groundtruth representation of Pavia University}
\end{figure}

The Pavia University scene is acquired by the 
Reflective Optics System Imaging Spectrometer (ROSIS) sensor. 
Removing the noisy bands and a blank strip, 
the data size in format 
$\mathsf{height}\times\mathsf{width}\times{\mathsf{depth}}$ is 
$610\,\mathsf{pixel}\times 340\,\mathsf{pixel}\times 103\,\mathsf{band}$. 
The spatial resolution is about $1.3$ meters. 
As shown in \cref{data_paviau}, the pixels are labeled with $9$ classes. 
In practice, $200\times 9$ labeled samples are 
chosen to form the training set, while the rest of the labeled samples 
form the testing set. The non-labeled pixels constitute the backgrounds. 
\cref{paviau_fc_gt} depicts the false color composite and the representation of groundtruth.
% pavia u table and image start 
\begin{table}[!htb]\caption{\label{data_paviau}Reference classes and sizes of training and testing sets of Pavia University image}
	\centering
	\begin{tabular}{ccccc}
		\hline
		\hline
		No. 	&Class	&Cardinality	&Train	&Test\\
		\hline
		$1$	&Asphalt	&$6631$	&$200$	&$6431$	\\
		$2$	&Meadows	&$18649$	&$200$	&$18449$	\\
		$3$	&Gravel	&$2099$	&$200$	&$1899$	\\
		$4$	&Trees	&$3064$	&$200$	&$2864$	\\
		$5$	&Painted	metal	sheets	&$1345$	&$200$	&$1145$	\\
		$6$	&Bare	Soil	&$5029$	&$200$	&$4829$	\\
		$7$	&Bitumen	&$1330$	&$200$	&$1130$	\\
		$8$	&Self-Blocking	Bricks	&$3682$	&$200$	&$3482$	\\
		$9$	&Shadows	&$947$	&$200$	&$747$	\\
		\hline
		&Total&$42776$&$1800$&$40976$	\\
		\hline
		\hline
	\end{tabular}
\end{table}
% pavia u table and image end

\begin{figure}[!htb]
	\centering
\graphicspath{{Figures/}}
	\includegraphics[trim =16mm 20mm 25mm 10mm, clip,width=.22\textwidth] {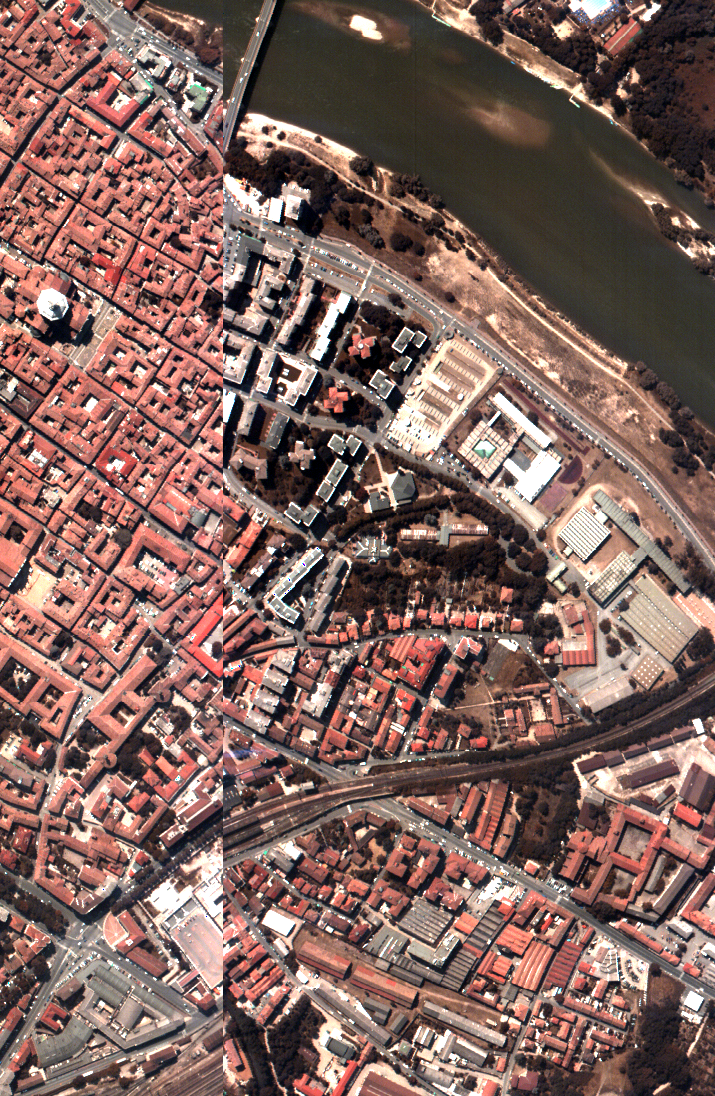}
	\includegraphics[trim =16mm 20mm 25mm 10mm, clip,width=.22\textwidth]{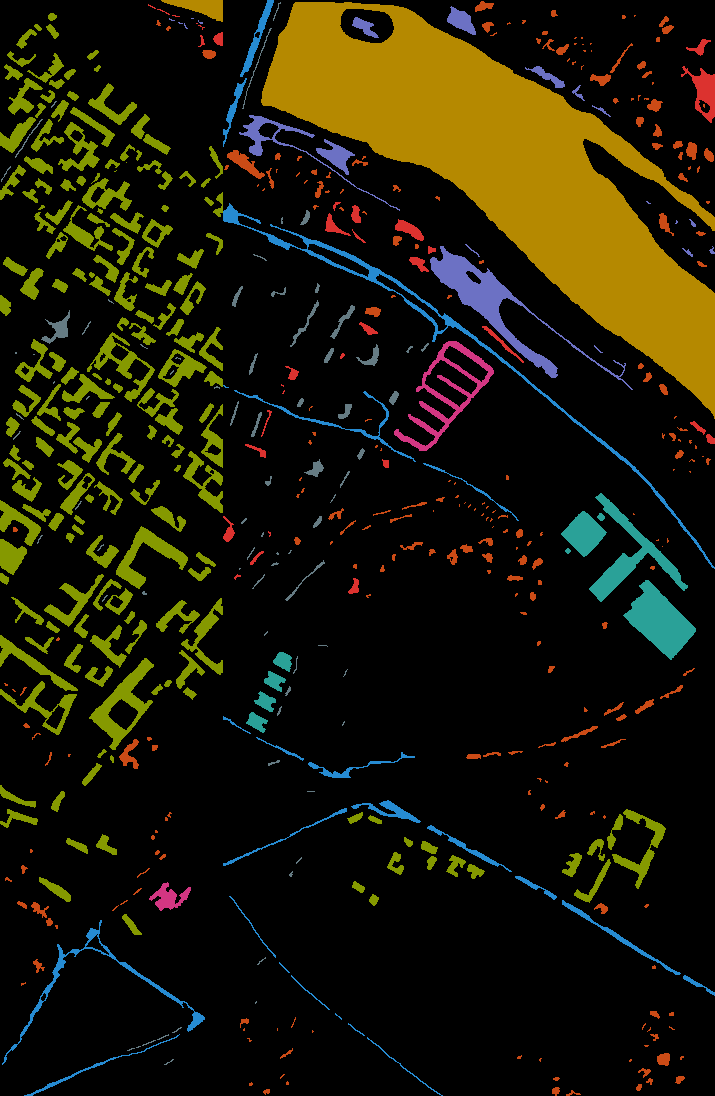}
	\includegraphics[trim =0mm 0mm 0mm 0mm, clip,width=.2\textwidth]{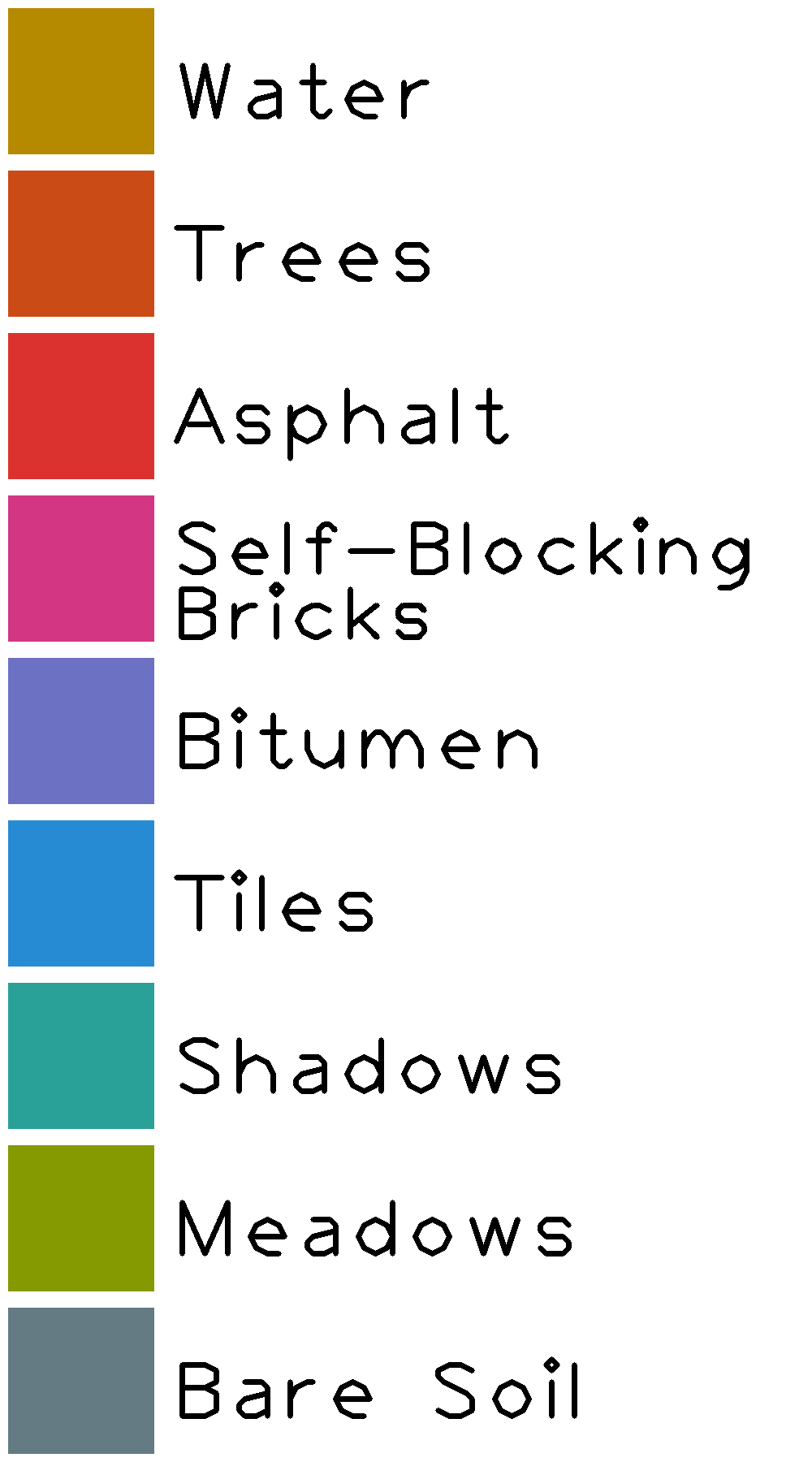}
	\caption{\label{paviac_fc_gt} The false color composite (band $10,20,40$) and groundtruth map of Pavia Centre} %{\green (10, 20, 40)}
\end{figure}

The Pavia Centre scene is also acquired by the 
ROSIS sensor. 
The data size is 
$1096\,\mathsf{pixel}\times 715\,\mathsf{pixel}\times 102\,\mathsf{band}$
after removing the noisy bands.
The spatial resolution is also $1.3$ meters. 
As illustrated in \cref{data_paviac}, the pixels in the Pavia University scene 
are labeled with $9$ classes. 
The constructions of the training and testing set are in the same way as 
in the Pavia University scene. 
\cref{paviac_fc_gt} illustrates the false color composite and the representation of groundtruth.
% pavia c table and image start 
\begin{table}[!htb]\caption{\label{data_paviac}Reference classes and sizes of training and testing sets of Pavia Centre image}
	\centering
	\begin{tabular}{ccccc}
		\hline
		\hline
		No. 	&Class	&Cardinality	&Train	&Test	\\
		\hline
		$1$	&Water	&$65971$	&$200$	&$65771$	\\
		$2$	&Trees	&$7598$	&$200$	&$7398$	\\
		$3$	&Asphalt	&$3090$	&$200$	&$2890$	\\
		$4$	&Self-Blocking	Bricks	&$2685$	&$200$	&$2485$	\\
		$5$	&Bitumen	&$6584$	&$200$	&$6384$	\\
		$6$	&Tiles	&$9248$	&$200$	&$9048$	\\
		$7$	&Shadows	&$7287$	&$200$	&$7087$	\\
		$8$	&Meadows	&$42826$	&$200$	&$42626$	\\
		$9$	&Bare	Soil	&$2863$	&$200$	&$2663$	\\
		\hline
		&Total&$148152$&$1800$&$146352$	\\
		\hline
		\hline
	\end{tabular}
\end{table}
% pavia c table and image end

\begin{figure}[!htb]
	\centering
\graphicspath{{Figures/}}
	\includegraphics[trim = 2mm 22mm 13mm 12mm, clip,width=.2\textwidth]  {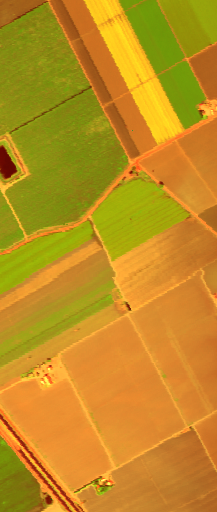}
	\includegraphics[trim =  2mm 22mm 13mm 12mm,clip,width=.2\textwidth]  {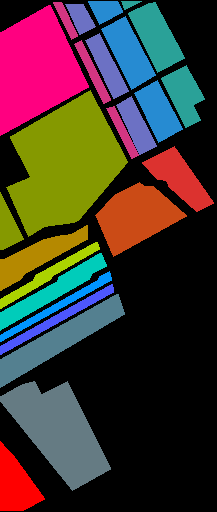}
	\includegraphics[trim =  0mm 0mm 0mm 0mm,clip,width=.21\textwidth]  {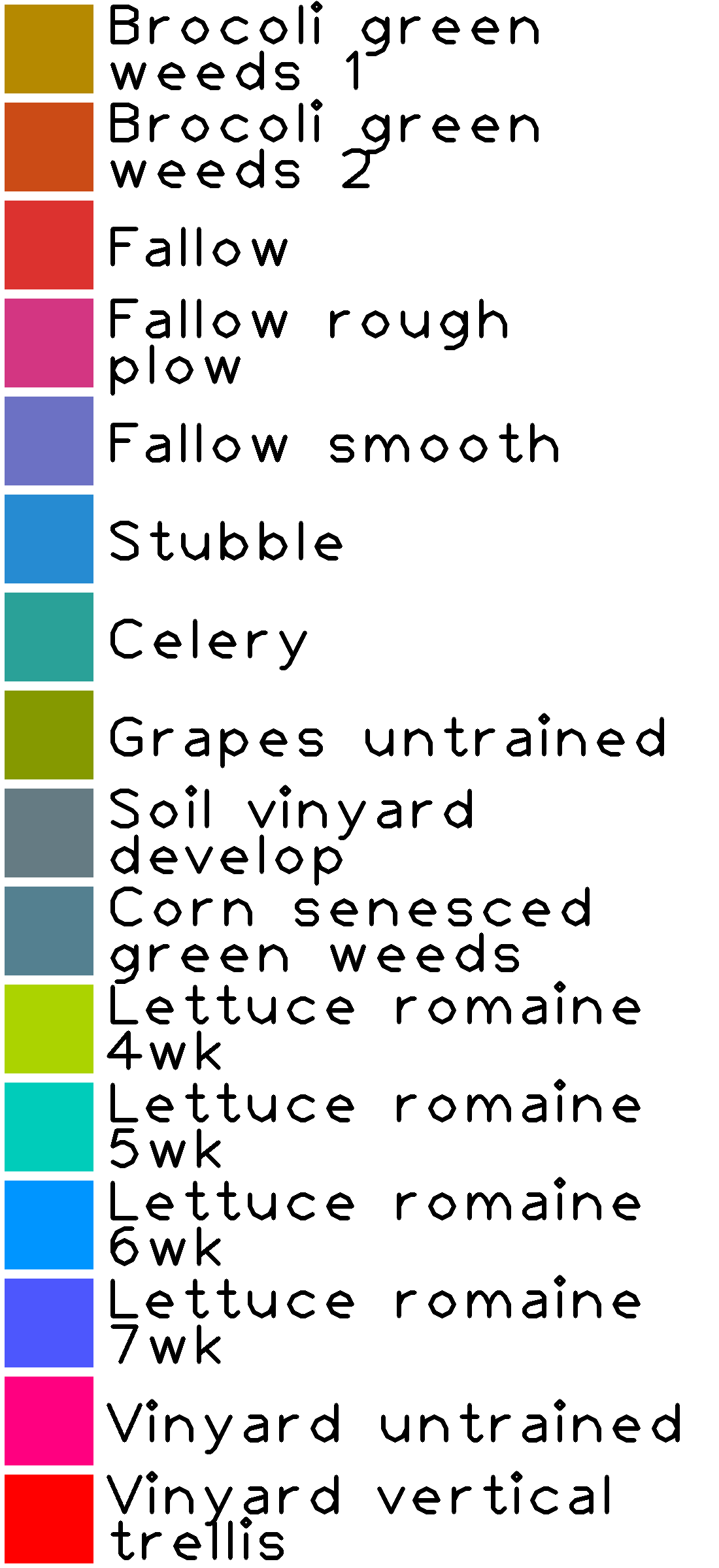}	
	\caption{\label{salinas_fc_gt} The false color composite (band $180,100,10$) and groundtruth representation of Salinas}
\end{figure}

The Salinas scene is 
collected by the Airborne Visible Infrared Imaging Spectrometer (AVIRIS). 
After the removal of the water absorption bands, the remained HSI has a size of
$512\,\mathsf{pixel}\times 217\,\mathsf{pixel}\times 204\,\mathsf{band}$. 
The pixels are classified into $16$ categories, 
and $200\times 16$ samples are picked for training, 
as shown in \cref{data_salinas}. 
The false color composite and the representation of groundtruth is shown 
in \cref{salinas_fc_gt}.
% salinas table and image start 
\begin{table}[!htb]\caption{\label{data_salinas}Reference classes and sizes of training and testing sets of Salinas image}
	\centering
	\begin{tabular}{ccccc}
		\hline
		\hline
		No. 	&Class	&Cardinality	&Train&   Test	\\
		\hline
		$1$	&Brocoli	green	weeds	1	&$2009$	&$200$	&$1809$	\\
		$2$	&Brocoli	green	weeds	2	&$3726$	&$200$	&$3526$	\\
		$3$	&Fallow	&$1976$	&$200$	&$1776$	\\
		$4$	&Fallow	rough	plow	&$1394$	&$200$	&$1194$	\\
		$5$	&Fallow	smooth	&$2678$	&$200$	&$2478$	\\
		$6$	&Stubble	&$3959$	&$200$	&$3759$	\\
		$7$	&Celery	&$3579$	&$200$	&$3379$	\\
		$8$	&Grapes	untrained	&$11271$	&$200$	&$11071$	\\
		$9$	&Soil	vinyard	develop	&$6203$	&$200$	&$6003$	\\
		$10$	&Corn	senesced	green	weeds	&$3278$	&$200$	&$3078$	\\
		$11$	&Lettuce	romaine	4wk	&$1068$	&$200$	&$868$	\\
		$12$	&Lettuce	romaine	5wk	&$1927$	&$200$	&$1727$	\\
		$13$	&Lettuce	romaine	6wk	&$916$	&$200$	&$716$	\\
		$14$	&Lettuce	romaine	7wk	&$1070$	&$200$	&$870$	\\
		$15$	&Vinyard	untrained	&$7268$	&$200$	&$7068$	\\
		$16$	&Vinyard	vertical	trellis	&$1807$	&$200$	&$1607$	\\
		\hline
		&Total&$54129$&$3200$&$50929$	\\
		\hline
		\hline
	\end{tabular}
\end{table}
% salinas table and image end

The Houston2018 (grss\_dfc\_2018) scene is acquired by 
the National Center for Airborne Laser Mapping 
over the University of Houston campus and its neighborhood. 
The size of this HSI is 
$601\,\mathsf{pixel}\times 2384\,\mathsf{pixel}\times 48\,\mathsf{band}$, with 
a $1$-meter ground sample distance. 
However, the groundtruth matrix has a quadrupled size 
$1202\,\mathsf{pixel}\times 4768\,\mathsf{pixel}$ with a $0.5$-meter ground sample 
distance. 
In practice, the label for each pixel is determined 
by the largest vote strategy using the groundtruth matrix. 
As illustrated in \cref{data_houston}, there are $20$ classes 
in the grss\_dfc\_2018, and a large variance exists among 
the sample numbers of different classes. 
For each class, the size of the training set is chosen as $20\%$ of the samples, 
truncated by 3200.

Finally, for all the datasets, the number of 
training samples in each class is enlarged to $3200$ by 
data augmentation, using rotation, mirroring and duplicating.

% houston table and image start 
\begin{table}[!htb]\caption{\label{data_houston}Reference classes and sizes of training and testing sets of grss\_dfc\_2018 image}
	\centering
	\begin{tabular}{ccccc}
		\hline \hline
		No. & Class                     & Cardinality & Train & Test   \\
		\hline
		1   & Healthy grass             & $9799$        & $1959$  & $7840$   \\
		2   & Stressed grass            & $32502$       & $3200$  & $29302$  \\
		3   & Artificial turf           & $684$         & $136$   & $548$    \\
		4   & Evergreen trees           & $13107$       & $2621$  & $10486$  \\
		5   & Deciduous trees           & $4810$        & $962$   & $3848$   \\
		6   & Bare earth                & $4516$        & $903$   & $3613$   \\
		7   & Water                     & $266$         & $53$    & $213$    \\
		8   & Residential buildings     & $38268$       & $3200$  & $35068$  \\
		9   & Non-residential buildings & $221145$      & $3200$  & $217945$ \\
		10  & Roads                     & $41178$       & $3200$  & $37978$  \\
		11  & Sidewalks                 & $28609$       & $3200$  & $25409$  \\
		12  & Crosswalks                & $1399$        & $279$   & $1120$   \\
		13  & Major thoroughfares       & $44933$       & $3200$  & $41733$  \\
		14  & Highways                  & $9507$        & $1901$  & $7606$   \\
		15  & Railways                  & $6937$        & $1387$  & $5550$   \\
		16  & Paved parking lots        & $10725$       & $2145$  & $8580$   \\
		17  & Unpaved parking lots      & $129$         & $25$    & $104$    \\
		18  & Cars                      & $4835$        & $967$   & $3868$   \\
		19  & Trains                    & $4622$        & $924$   & $3698$   \\
		20  & Stadium seats             & $6824$        & $1364$  & $5460$   \\
		\hline
			& Total       & $484795$      & $34826$ & $449969$ \\
		\hline \hline
	\end{tabular}
\end{table}
\begin{figure}[!htb]
	\centering
\graphicspath{{Figures/}}
	\includegraphics[trim = -10mm -10mm -10mm -10mm, clip,width=0.67\textwidth]  {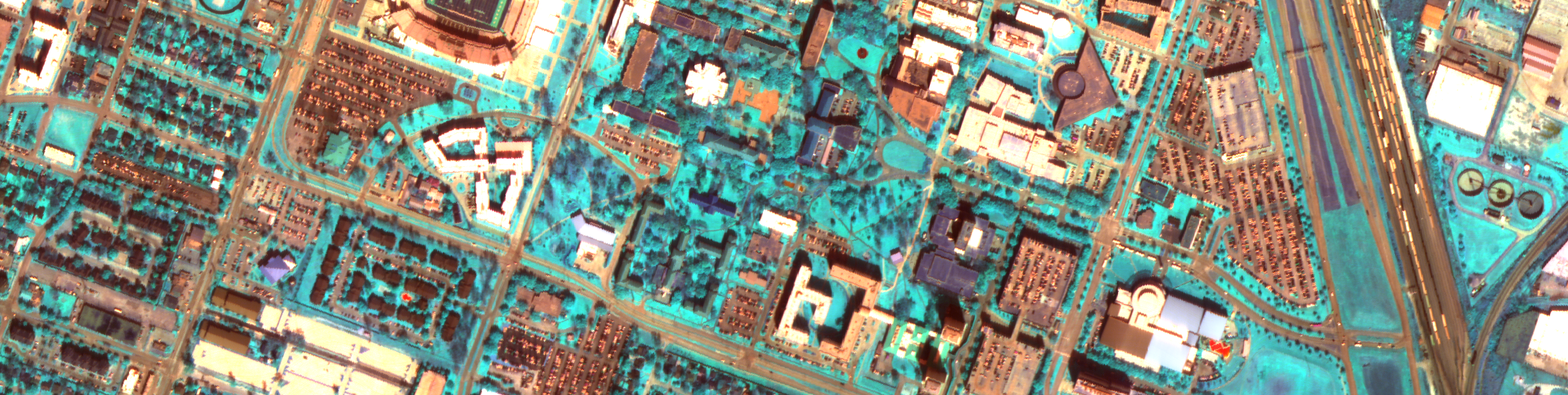}
	\includegraphics[trim = -10mm -10mm -10mm -10mm,clip,width=0.67\textwidth]  {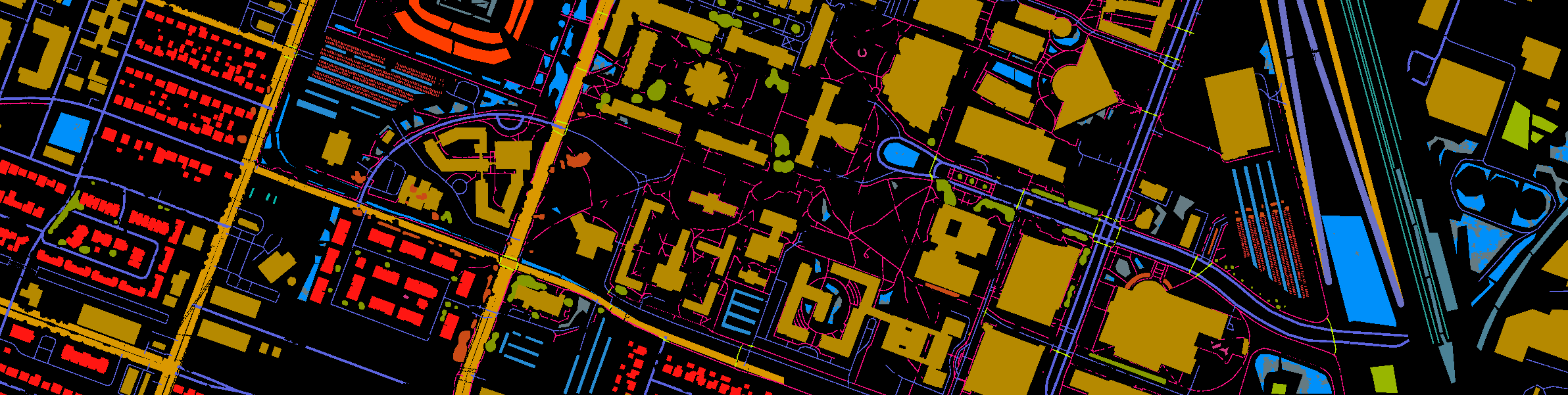}
	\includegraphics[trim = -10mm -10mm -10mm -10mm,clip,width=0.67\textwidth]  {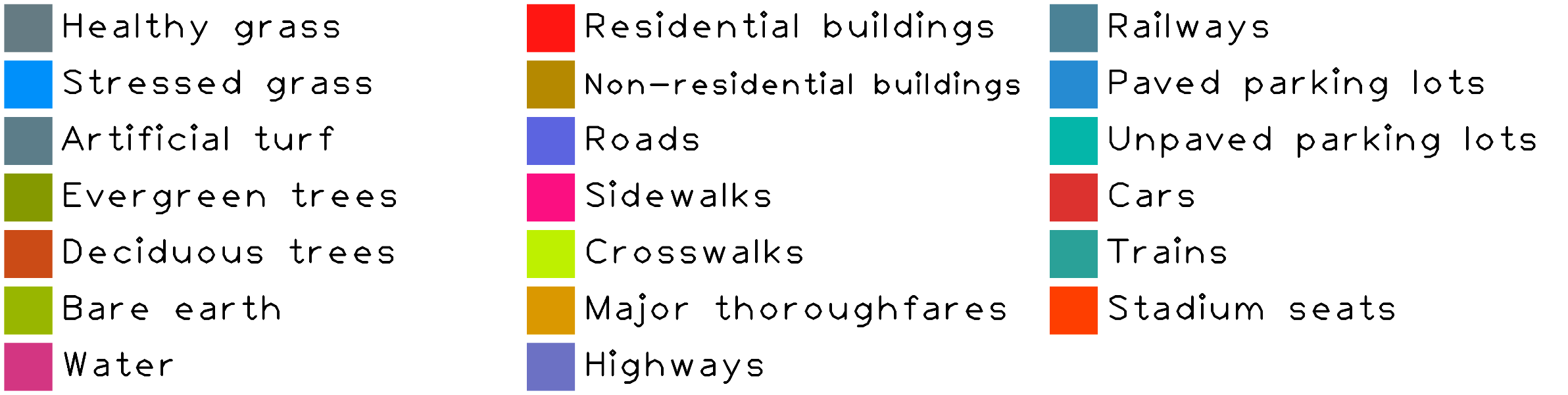}	
	\caption{\label{houston_fc_gt} The false color composite (band $48,28,8$) and groundtruth representation of grss\_dfc\_2018}
\end{figure}

\subsection{Ablation study and comparative experiments}
We design the ablation study to verify that the effectiveness 
of the proposed ABMHC is mainly attributed to the following two factors: 
1) The abundance features extracted by autoencoder-based SU 
have more discriminative ability than the raw spectra, 
hence they are better classified by the CNN-based classifier; 
2) The combination of abundance representations 
from different HSIs yields a compatible classifier 
that is more powerful than the data-specific classifier. 
As a baseline, we train a CNN-based classifier 
directly on the raw spectral data for each HSI, 
and term this method as raw-CNN. 
Besides, 
the abundance-based HSI classification is performed on each dataset, 
and we refer this series of experiments as abun-CNN. 
The first aforementioned factor can be evaluated by comparing the 
results of raw-CNN and abun-CNN. 
Finally, 
we perform the abundance-based 
and multi-HSIs-based algorithm on the merged big training set, which 
is the proposed ABMHC.
To conduct the ablation study, the SU procedure is set to be the same for the abun-CNN and the ABMHC, namely for every HSI scene, same endmember matrix and abundance representation are considered in both methods. 
The only difference between the abun-CNN and the ABMHC is whether the classifier is involved with multi-HSIs.\footnote{To train a compatible classifier 
directly using the collection of raw spectra from different datasets 
cannot be realized, as the number of bands varies in different HSIs.}
To keep a fair comparison, the same network structures and hyperparameters 
are utilized in raw-CNN, abun-CNN and ABMHC, as explicated 
in \cref{sec:autoencoder} and \cref{sec:cnn}. 
In the experiments of abun-CNN and ABMHC, the parts of SU are identity. 
The size of abundance patches and HSI patches adopted in the CNNs
of the proposed ABMHC and its ablation study is set to  
$11\, \mathsf{pixel}\times 11\, \mathsf{pixel}$. 
The effect of this parameter on classification 
performance is not analyzed in this paper, 
as investigations have already been made in several existing works
\cite{guo2019spectral,cao2019spectral}.

To further evaluate the performance of proposed ABMHC, 
we choose three lately-proposed classification algorithms 
for comparison, namely the method of 
pixel-pair feature (PPF)\cite{li2017hyperspectral},
the spectral-spatial squeeze-and-excitation residual bag-of-feature learning (S3EResBoF)\cite{roy2020lightweight}, 
and the hybrid spectral net (HybridSN)\cite{roy2019hybridsn}.
All these state-of-the-art methods are based on deep learning with CNN and performed on raw spectra of HSIs.
They have shown promising classification results on several HSIs datasets.
For fairness, all the comparing methods are performed using 
the training sets with the same size, as described in~\cref{PrepareTraining},
except for the PPF algorithm on grss\_dfc\_2018. 
In fact, the PPF generates pixel pairs as training samples, 
so that the size of the training set is squared. 
This leads to an out-of-memory situation on our
server equipped with $256$G RAM. 
In practice, the original training set for PPF on 
grss\_dfc\_2018 is constructed by choosing $20\%$ 
of the labeled samples from each class, 
and truncating the number by $1600$. 
For this reason, the performance of PPF on this dataset is not satisfactory, 
as to be given in~\cref{results_houston}.
To roughly explore the effects 
of introducing abundance representation on complex structured networks, we also perform the HybridSN 
using the abundance features extracted by autoencoder-based SU on all the four HSI datasets.

\subsection{Results analysis}
We apply three commonly used metrics to evaluate the 
performances of all the algorithms, namely OA, AA, and $\kappa$. 
The overall accuracy (OA) represents the ratio of the correctly 
classified samples number to the total  samples number; 
the average accuracy (AA) is the mean accuracy of different classes; 
the Cohen's kappa coefficient $\kappa$ measures the agreement between 
the predicted labels and the groundtruth labels. 

\begin{table}[!htb]\caption{\label{results_paviau}
	classification accuracies (averaged over 5 runs) on Pavia University scene}
	\centering
		\begin{tabular}{cccc|ccc}
			\hline \hline
		No.                    & raw-CNN & abun-CNN & ABMHC & PPF    &	S3EResBoF& HybridSN(raw/abun)\\
		\hline
		1              & 94.59   & 96.45   & 97.99 & 97.25  &	99.35& \textbf{99.74}/99.67\\
		2              & 95.90   & 98.12   & 98.87 & 95.24  &	\textbf{99.88}& 99.43/99.63\\
		3               & 92.70   & 96.46   & \textbf{98.87} & 94.17  &	98.79& 97.22/97.22\\
		4                & 98.65   & 98.58   & \textbf{98.86} & 97.20  &	88.04& 96.90/96.24\\
		5 & \textbf{100.00}  & 99.90   & 99.98 & \textbf{100.00} &	99.64& 99.95/98.84\\
		6            & 94.54   & 99.34   & \textbf{99.40} & 99.37  &	99.02& 98.50/98.21\\
		7              & 97.86   & 98.69   & \textbf{99.00} & 96.16  &	98.87& 95.66/97.60		\\
		8 & 93.76   & 95.39   & \textbf{98.67} & 93.83  &	96.25& 97.36/97.84\\
		9              & 99.68   & 99.65   & \textbf{99.90} & 99.46  &	96.22& 99.26/99.21\\
		\hline
		OA$(\%)$             & 95.63   & 97.82   & 98.83 & 96.25  &	97.65& 98.81/\textbf{98.87}\\
		AA$(\%)$             & 96.41   & 98.06   & \textbf{99.06} &96.97  & 	97.34& 98.23/98.27\\
		$\kappa(\times 100)$ & 94.18   & 97.07   & 98.43 & 94.99  &	96.92& 98.39/\textbf{98.48}\\
		\hline \hline
		\end{tabular}
\end{table}

\begin{table}[!htb]\caption{\label{results_paviac}
	classification accuracies (averaged over 5 runs) on Pavia Centre scene}
	\centering
	\begin{tabular}{cccc|ccc}
		\hline \hline
		No.				& raw-CNN & abun-CNN & ABMHC & PPF   &	S3EResBoF&  HybridSN(raw/abun) \\
		\hline
		1                & 99.89   & 99.83   & 99.95 & 99.15 &	\textbf{100.00}&  \textbf{100.00}/99.99   \\
		2                & 95.32   & 97.29   & 97.61 & 97.96 &	99.14&  \textbf{99.73}/99.49    \\
		3              & 96.73   & 96.81   & \textbf{97.73} & 97.37 &	91.69&  94.20/95.34    \\
		4 & 98.74   & 99.78   & \textbf{99.86} & 99.27 &	99.81&  96.48/97.54    \\
		5              & 98.62   & 98.58   & 99.17 & 98.79 &	98.82&  99.24/\textbf{99.64}    \\
		6                & 98.36   & 98.47   & 98.90 & 98.95 &	\textbf{99.13}&  98.39/98.43    \\
		7              & 96.40   & 97.44   & 98.82 & 94.36 &	\textbf{99.97}&  99.23/99.44    \\
		8              & 99.66   & 99.80   & 99.84 & 99.90 &	99.87&  99.97/\textbf{99.99}    \\
		9            & 99.87   & 99.86   & \textbf{99.96} & \textbf{99.96} &	91.63&  95.92/97.79    \\
		\hline
		OA$(\%)$             & 99.19   & 99.38   & 99.60 & 99.03 &	99.45&  99.55/\textbf{99.65}    \\
		AA$(\%)$             & 98.18   & 98.65   & \textbf{99.09} & 98.41 &	97.78&  98.13/98.63    \\
		$\kappa(\times 100)$ & 98.85   & 99.12   & 99.43 & 98.62 &	99.22&  99.35/\textbf{99.49}    \\
		\hline \hline
		\end{tabular}
\end{table}

\begin{table}[!htb]\caption{\label{results_salinas}
	classification accuracies (averaged over 5 runs) on Salinas scene}
	\centering
	\begin{tabular}{cccc|ccc}
		\hline \hline
		No.					   & raw-CNN & abun-CNN & ABMHC  & PPF    & 	S3EResBoF& HybridSN(raw/abun) \\
		\hline
		1         & 99.76   & 99.96   & \textbf{100.00} & 99.98  & 	99.00& 99.98/\textbf{100.00}    \\
		2         & 99.17   & 99.80   & 99.72  & 99.58  & 	\textbf{100.00}& 99.88/99.93    \\
		3                        & 98.41   & 99.64   & 99.79  & 99.61  & 	\textbf{99.97}& 99.76/99.76    \\
		4             & 99.16   & 99.35   & 99.69  & \textbf{99.73}  & 	99.68& 98.67/99.44    \\
		5                 & 97.75   & 99.11   & 99.56  & 97.43  & 	\textbf{99.90}& 99.83/99.43    \\
		6                       & 99.58   & 99.70   & 99.96  & 99.66  &  	\textbf{100.00}&99.92/\textbf{100.00}    \\
		7                        & 99.63   & 99.76   & 99.82  & \textbf{99.93}  & 	99.87& 99.67/\textbf{99.93}    \\
		8              & 81.68   & 88.93   & 94.21  & 84.81  & \textbf{98.66}& 96.67/97.15    \\
		9 & 97.15   & 99.04   & 99.68  & 99.15  & 	\textbf{100.00}& 99.64/99.94    \\
		10     & 93.24   & 97.93   & 98.16  & 96.73  & 	\textbf{99.61}& 99.00/98.83    \\
		11           & 99.59   & 99.31   & 99.60  & 99.45  & 	\textbf{99.70}& 98.58/99.43    \\
		12           & 99.81   & 99.97   & \textbf{100.00} & \textbf{100.00} & 	99.99& 99.88/99.93    \\
		13           & 99.61   & \textbf{100.00}  & 99.97  & 99.50  & 	\textbf{100.00}& \textbf{100.00}/99.94   \\
		14           & 99.63   & 99.98   & \textbf{100.00} & 99.47  & 	99.90& 99.59/99.84    \\
		15             & 84.26   & 91.66   & \textbf{95.40}  & 81.80  & 	95.05& 92.22/90.07    \\
		16      & 99.20   & 99.18   & 99.27  & 98.81  & 	\textbf{99.99}& 99.07/99.53    \\
		\hline
		OA$(\%)$                      & 92.73   & 96.03   & 97.85  & 93.61  & 	\textbf{98.91}& 97.91/97.73    \\
		AA$(\%)$                      & 96.73   & 98.33   & 99.05  & 97.23  & 	\textbf{99.46}& 98.90/98.95    \\
		$\kappa(\times 100)$          & 91.89   & 95.56   & 97.60  & 92.85  & 	\textbf{98.78}& 97.66/97.46    \\
		\hline \hline
	\end{tabular}
\end{table}

\begin{table}[!htb]\caption{\label{results_houston}
	classification accuracies (averaged over 5 runs) on grss\_dfc\_2018 scene}
	\centering
	\begin{tabular}{cccc|ccc}
		\hline \hline
						   & raw-CNN & abun-CNN & ABMHC & PPF            & 	S3EResBoF&HybridSN(raw/abun)\\
		\hline
		1             & 96.63   & \textbf{96.83}   & 96.03 & \textit{87.18} &	81.38& 88.56/79.92\\
		2            & 93.76   & 94.32   & 95.54 & \textit{94.04} &	\textbf{96.79}& 88.56/96.63\\
		3           & 99.01   & 99.67   & \textbf{99.86} & \textit{98.92} &	99.80& 88.56/95.41\\
		4           & 98.47   & 98.77   & \textbf{98.78} & \textit{84.94} &	92.80& 90.17/91.88\\
		5           & 95.95   & \textbf{97.16}   & 96.23 & \textit{55.80} &	86.70& 84.51/80.86\\
		6                & 99.22   & \textbf{99.81}   & 99.66 & \textit{93.81} &	99.33& 96.92/97.19\\
		7                     & 96.06   & 92.02   & 98.47 & \textit{99.50} &	\textbf{99.19}& 91.16/87.65\\
		8     & 90.98   & 93.49   & \textbf{94.72} & \textit{68.71} &	91.96& 85.79/88.70\\
		9 & 86.52   & 90.53   & 92.81 & \textit{98.26} &	\textbf{99.72}& 99.28/99.03\\
		10                     & 65.32   & 70.77   & 75.32 & \textit{71.75} &	\textbf{91.42}& 85.09/83.71\\
		11                 & 72.27   & 78.62   & \textbf{82.89} & \textit{59.13} &	78.54& 71.22/70.79\\
		12                & 60.20   & 71.93   & \textbf{77.83} &\textit{39.42} &	43.78&  37.02/42.72\\
		13       & 75.94   & 80.78   & 87.19 & \textit{78.71} &	\textbf{95.12}& 91.21/91.52\\
		14                  & 98.35   & \textbf{98.78}   & 98.63 & \textit{55.89} &	95.26& 89.96/87.98\\
		15                  & 99.33   & 99.78   & \textbf{99.88} & \textit{93.24} &	99.49& 97.80/97.88\\
		16        & 96.49   & 97.34   & 97.51 & \textit{85.49} & 	\textbf{98.66}& 94.12/93.83\\
		17      & 96.73   & 98.27   & 98.08 & \textit{98.13} &	\textbf{100.00}& 81.07/89.14\\
		18                      & 94.61   & \textbf{96.20}   & 95.91 & \textit{37.25} &	91.82& 67.72/65.30\\
		19                    & 98.29   & \textbf{99.23}   & 99.02 &\textit{61.27} &	98.69&  88.18/93.55\\
		20             & 99.57   & 99.82   & \textbf{99.85} &\textit{73.22} &	97.95&  88.03/91.15\\
		\hline
		OA$(\%)$                  & 85.23   & 88.78   & 91.36 & \textit{83.47} &	\textbf{95.14}& 91.86/92.12\\
		AA$(\%)$                  & 90.68   & 92.71   & \textbf{94.21} & \textit{76.73} &	91.92& 85.14/86.24\\
		$\kappa(\times 100)$      & 80.60   & 85.10   & 88.79 & \textit{78.51} &	\textbf{93.47}& 89.13/89.45\\
		\hline \hline
		\end{tabular}
\end{table}

\begin{table}[!htb]
\caption{\label{results_compara}
classification accuracies on Pavia University and 
Pavia Centre, with rearranged classes order}
	\centering
	\begin{tabular}{ccc|c|ccc|c}
		\hline \hline
		\multicolumn{4}{c|}{Pavia University}               & \multicolumn{4}{c}{Pavia Centre}                   \\\hline
		Class                & abun-CNN & ABMHC & Progress & Class                & abun-CNN & ABMHC & Progress \\\hline
		Asphalt              & 96.45    & 97.99 & 1.54     & Asphalt              & 96.81    & 97.73 & 0.92     \\
		Meadows              & 98.12    & 98.87 & 0.75     & Meadows              & 99.80    & 99.84 & 0.04     \\
		Trees                & 98.58    & 98.86 & 0.28     & Trees                & 97.29    & 97.61 & 0.32     \\
		Bare Soil            & 99.34    & 99.40 & 0.06     & Bare Soil            & 99.86    & 99.96 & 0.1      \\
		Bitumen              & 98.69    & 99.00 & 0.31     & Bitumen              & 98.58    & 99.17 & 0.59     \\
		SBB & 95.39    & 98.67 & 3.28     				   & SBB & 99.78    & 99.86 & 0.08     \\
		Shadows              & 99.65    & 99.90 & 0.25     & Shadows              & 97.44    & 98.82 & 1.38     \\ \hline
		\textsl{PMS} & 99.90    & 99.98 & 0.08     & \textsl{Water}                & 99.83    & 99.95 & 0.12     \\
		\textsl{Gravel}               & 96.46    & 98.87 & 2.41     & \textsl{Tiles}                & 98.47    & 98.90 & 0.43    \\
		\hline \hline
	\end{tabular}
\end{table}

The results obtained by the proposed ABMHC, the ablation study methods, 
{\em i.e.}, raw-CNN and abun-CNN and three state-of-the-art  
methods on the aforementioned HSI datasets are listed in  
\cref{results_paviau,results_paviac,results_salinas,results_houston}\footnote{The results of PPF on grss\_dfc\_2018 
are in italic, 
as the number of training samples used in this case 
is not the same as in other methods.}. 

We observe the following facts from the ablation study.
Firstly, the abun-CNN method always outperforms the raw-CNN method with 
a large margin in terms of all the metrics, on all the datasets. 
This indicates that compared with the raw spectral features, 
the abundance features extracted by autoencoder demonstrate 
improved discriminative ability, thus improving 
the performance of the classifier. 
Secondly, compared with the abun-CNN method, 
which employs the abundance representation from one single HSI, 
the multi-HSI based ABMHC always leads to better classification 
performances in all the metrics, on all the datasets. 
This demonstrates that enlarging the training set by merging the 
abundance information from different HSIs augments the classifier performance. 
A convincing explanation of this phenomenon is that the 
merged training set alleviates the overfitting issue on the 
CNN-based classifier by increasing the number of training samples. 

In practice, different HSI scenes may have several classes in common, {\em e.g.}, 
Pavia University and Pavia Centre both have 9 classes, 7 of them being the same.  
However, such coincidence of classes in different datasets will not affect the 
effectiveness of the proposed method, as to be explained in following aspects.
Firstly, even for the HSIs with mutual classes, the chance that 
the elements in abundance features from different HSIs correspond to the same endmembers, 
in the same order, is very slim. It is because the permutation of endmembers order 
in $\hat{\bm{M}}$ is not taken into account for each data. 
Secondly, the assemblage of abundance representations from different HSIs is only used to train the classifier,
while the testing stage is performed on each specific dataset. 
As a result, the testing stage will not be affected by the existence of the shared classes in different HSIs. 
To show the influence of the shared classes on the training stage, we have rearranged the results of Pavia University and 
Pavia Centre, two datasets with great coincidence of classes, in \cref{results_compara}. 
Compared to the abun-CNN, the proposed fusion technique always leads to superior results on every class, 
whether it appears in both images or not. This demonstrates that the coincidence of classes 
has little negative effects on the performance. 

The proposed ABMHC  
leads to comparable performance to the state-of-the-art algorithms. 
It outperforms the PPF method with a large margin 
on all the datasets. 
In comparison with the latest S3EResBoF and HybridSN, 
the proposed ABMHC generally provides comparable results. 
To be precise, our method slightly outperforms S3EResBoF and HybridSN on 
the Pavia University scene and 
the Pavia Centre scene. 
On the Salinas scene, the results of ABMHC fall between the results of the HybridSN and S3EResBoF.
On grss\_dfc\_2018 dataset, the proposed method surpasses the S3EResBoF and HybridSN by 
a large margin in AA, while inferior to its counterparts in OA and $\kappa$. 
It is worth noting that the proposed ABMHC utilizes a simple CNN 
classifier, which is 
plain and shallow, while the S3EResBoF and the HybridSN employ far more complicated network 
structures \cite{roy2020lightweight, roy2019hybridsn}. 

The effects 
of introducing abundance representation to the networks with complex structure are shown in the 
the last columns of \cref{results_paviau,results_paviac,results_salinas,results_houston}, 
where the HybridSN is conducted using both the raw spectral data and the abundance representations. 
Compared to the raw-data-based counterpart, classification with the abundance representations 
leads to slight improvements on all the datasets.
This demonstrates that even for the complex networks, the abundance strategy still works. 

\section{Conclusion}\label{sec:conclusion}
In this paper, we proposed an abundance-based multi-HSI classification method, to address the overfitting issue in deep learning-based classification.  
The original intention of the proposed method is two-fold: 
1) The abundance features extracted by SU have more discriminative 
ability than the raw spectral features, which enables the use of 
simple networks to alleviate the overfitting issue; 
2) Training a classifier with multiple HSIs will lead to 
superior performance than training with one single HSI, 
as enlarging the training set usually alleviates the 
overfitting issue. 
This idea becomes feasible by transforming multiple HSIs 
from the spectral domain to the abundance domain by SU. 

From these two aspects, we first designed and trained autoencoder-based SU model 
for each HSI seperately. 
After that, the HSIs were mapped to the abundance domain by the learned autoencoders. 
Lastly, a compatible classifier was trained by the abundance features from multiple HSIs, and further applied to predict the labels on the testing sets. 
The ablation study and comparative experiments were performed on four datasets. 
The results in the ablation study confirmed the original intention.  
The comparative experiments showed that our method provided comparable classification performance 
to the state-of-the-art methods, but using a far more simplified model structure. 

\section*{Acknowledgment}
This work was supported by the National Natural Science Foundation of China under Grant 11801409, 61701337, and the Natural Science Foundation of Tianjin City under Grand 18JCQNJC01600.

The authors would like to thank the National Center for 
Airborne Laser Mapping and the Hyperspectral Image Analysis 
Laboratory at the University of Houston for acquiring and 
providing the data used in this study, and the IEEE GRSS Image 
Analysis and Data Fusion Technical Committee.

  \bibliographystyle{elsarticle-num}
  \bibliography{AlanGuo,bibfei}


%\begin{thebibliography}{00}
%%% Text of bibliographic item
%
%\bibitem{}
%
%\end{thebibliography}
\end{document}